\documentclass[lettersize,journal]{IEEEtran}
\usepackage{amsmath,amsfonts}
\usepackage{algorithmic}
\usepackage{algorithm}
\usepackage{array}
\usepackage[caption=false,font=normalsize,labelfont=sf,textfont=sf]{subfig}
\usepackage{textcomp}
\usepackage{stfloats}
\usepackage{url}
\usepackage{verbatim}
\usepackage{graphicx}
\usepackage{cite}
\usepackage{algorithm}
\usepackage{algorithmic}
\usepackage{amsmath}
\usepackage{bm}
\usepackage{amssymb}
\usepackage{multirow}
\usepackage{array} 
\usepackage{makecell}
\usepackage{booktabs}

\begin{document}

\title{Transforming Noise Distributions with Histogram Matching: Towards a Single Denoiser for All}

\author{Sheng Fu, Junchao Zhang, \IEEEmembership{Member,~IEEE},  and Kailun Yang
        
\thanks{This work was supported in part by the National Natural Science Foundation of China under Grant No. 62105372, the Fundamental Research Foundation of National Key Laboratory of Automatic Target Recognition under Grant Number: WDZC20255290209, Hunan Provincial Research and Development Project under Grant No. 2025QK3019. \textit{(Corresponding author: Junchao Zhang.)}

Sheng Fu and Junchao Zhang are with Hunan Provincial Key Laboratory of Optic-Electronic Intelligent Measurement and Control, School of Automation, Central South University, Changsha 410083, China (e-mail: shengfu@csu.edu.cn; junchaozhang@csu.edu.cn).

Kailun Yang is with the School of Artificial Intelligence and Robotics, Hunan University, Changsha 410012, China (e-mail: kailun.yang@hnu.edu.cn).
}
}

\markboth{Journal of \LaTeX\ Class Files,~Vol.~14, No.~8, August~2021}%
{Shell \MakeLowercase{\textit{et al.}}: A Sample Article Using IEEEtran.cls for IEEE Journals}


\maketitle

\begin{abstract}
Supervised Gaussian denoisers exhibit limited generalization when confronted with out-of-distribution noise, due to the diverse distributional characteristics of different noise types. To bridge this gap, we propose a histogram matching approach that transforms arbitrary noise towards a target Gaussian distribution with known intensity. Moreover, a mutually reinforcing cycle is established between noise transformation and subsequent denoising. This cycle progressively refines the noise to be converted, making it approximate the real noise, thereby enhancing the noise transformation effect and further improving the denoising performance. We tackle specific noise complexities: local histogram matching handles signal-dependent noise, intrapatch permutation processes channel-related noise, and frequency-domain histogram matching coupled with pixel-shuffle down-sampling breaks spatial correlation. By applying these transformations, a single Gaussian denoiser gains remarkable capability to handle various out-of-distribution noises, including synthetic noises such as Poisson, salt-and-pepper and repeating pattern noises, as well as complex real-world noises. Extensive experiments demonstrate the superior generalization and effectiveness of our method. 
\end{abstract}

\begin{IEEEkeywords}
Image denoising, Histogram matching, Noise transformation.
\end{IEEEkeywords}

\section{Introduction}
\IEEEPARstart{I}{mages} are inevitably affected by various noises during the shooting process, leading to the destruction of original information. As a classic and important task in image processing, image denoising aims to recover clean images from noisy ones. It can not only improve image quality and visual effect, but also provide reliable data for high-level visual tasks such as object detection and semantic segmentation.

Traditional denoising methods such as CBM3D \cite{CBM3D} and low-rank denoising \cite{low-rank} utilize image prior information for denoising, with limited performance. In recent years, learning-based denoising methods have developed extensively and become mainstream. Among them, supervised learning-based denoising methods like DnCNN \cite{DnCNN} and Restormer \cite{Restormer} achieve excellent performance by training on large amounts of paired noisy and clean images. However, supervised methods require clean images, which may be unavailable in some scenarios. Thus, self-supervised denoising methods \cite{NB2NB,AP-BSN,TBSN} have emerged. They only use noisy images for training without needing clean ones, yet yield decent denoising results. Nevertheless, self-supervised denoising methods still require a large number of noisy images for training, prompting the development of zero-shot denoising methods \cite{Zs-n2n,S2S}. These methods only need a single noisy image for training, enabling adaptive denoising with good generalization.

\begin{figure}[t]
\centering
\includegraphics[width=0.94\columnwidth]{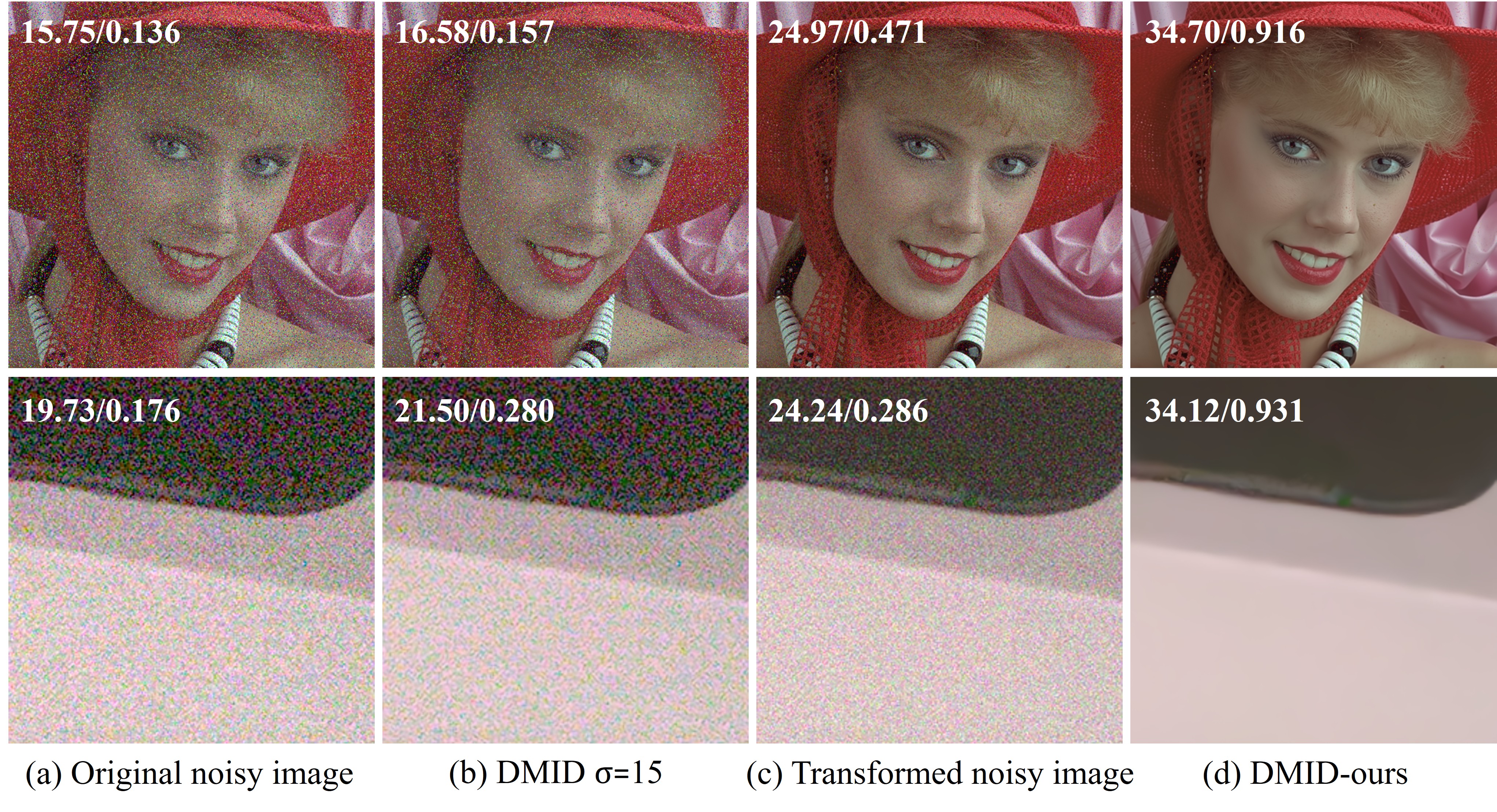}
\caption{Visualization of our noise transformation effect. The first row is random impulse noise and the second row is real-world noise. DMID \cite{DMID} is a diffusion-based powerful Gaussian denoiser, but it shows difficulty in removing out-of-distribution noise. Through our noise transformation, its denoising effect has been greatly improved.}
\label{fig1}
\end{figure}

However, despite the excellent denoising performance of existing learning-based methods, the generalization ability of models remains a problem. Both supervised and self-supervised denoising methods often fail to effectively remove new types of noise and require retraining. Although zero-shot denoising methods have better generalization, models trained on a single image hardly achieve good denoising performance. The generalization ability of Gaussian denoisers is more important, as many denoisers are designed for Gaussian noise. In fact, Gaussian denoisers are not limited to removing Gaussian noise, but they still cannot eliminate many out-of-distribution noises, such as periodic noise. Therefore, many methods have been proposed to improve the generalization ability of Gaussian denoisers. These methods can be divided into two categories. One starts from the perspective of noise, attempting to transform the noise so that it can be removed by Gaussian denoisers. For example, variance-stabilizing transformation \cite{VST} can convert Gaussian-Poisson noise to Gaussian noise. But it is not applicable to other types of noise and requires noise parameter estimation. The other starts from the perspective of model, changing the training data, network structure, etc. to make the model more robust. For example, Clipdenoising \cite{Clipde} incorporates a pretrained visual language model into its network structure to enhance generalization. While these methods do contribute to enhancing the generalization ability of Gaussian denoisers, they still struggle to handle many various out-of-distribution noises. 

In this paper, starting from the histogram distribution of noise, we propose a histogram matching method to address the generalization issue of Gaussian denoisers. Generally, the type and magnitude of noise can be well reflected in its histogram distribution, with different noises typically having distinct distributions. Given this, can one distribution be transformed into another? Histogram matching can achieve this. Thus, our motivation is to use histogram matching to convert out-of-distribution noise into in-distribution Gaussian noise, a straightforward idea. It is worth mentioning that histogram matching can effectively change the distribution of noise, and is thus very suitable to transform the noise. However, to our knowledge, there is currently no method for applying it to noise transformation. This is likely due to the emergence of a critical problem: how can we acquire such unknown out-of-distribution noise? It seems contradictory: transforming noise requires the noise itself, but the purpose of noise transformation is denoising. Therefore, if the noise is already known, there is no need for the transformation. Fortunately, although we cannot accurately obtain this unknown noise, we can get a rough estimate via image smoothing, which is much simpler than image denoising. Performing histogram matching on this estimated noise can also achieve noise transformation. But the effect of such noise transformation is limited. To achieve better effect, we establish a cycle between transformation and denoising. Specifically, we denoise the transformed noisy image, and then use the denoised result to obtain a more accurate noise estimate for the next round of transformation. This way, we realize mutual promotion between noise transformation and denoising. However, although histogram matching can modify the distribution of noise, it struggles to alter the noise’s spatial or channel correlation. Thus, it is necessary to introduce some shuffling methods. Inspired by \cite{PD} and \cite{Inrapatch}, we use Pixel-shuffle Down-sampling (PD) and intrapatch permutation to break the spatial and channel correlations.

Our method significantly enhances the generalization ability of Gaussian denoising models. As shown in Fig. \ref{fig1}, DMID \cite{DMID} originally performs poorly on random impulsive noise and real-world noise, but its denoising performance is greatly improved with our method. In our subsequent experiments, compared with DMID $\sigma=15$ (pre-transformation denoising), the PSNR and SSIM value of DMID-ours (post-transformation denoising) increased by 11.81dB and 0.517 respectively on various out-of-distribution noises. Finally, we summarize our contributions as follows:
\begin{itemize}
    \item We introduce the histogram matching method into noise transformation, enabling the conversion of noise with arbitrary distributions into Gaussian noise.
    \item We establish a cycle between noise transformation and denoising, realizing their mutual promotion.
    \item Extensive experiments demonstrate that our method endows existing Gaussian denoisers with the capability to remove various out-of-distribution noises, significantly enhancing their denoising performance.
\end{itemize}

The rest of this paper is organized as follows: Section II introduces the previously related works, inculding learning-based denoising methods and approaches that help improve the generalization of denoisers; Section III describes the proposed noise transformation method, detailing each step of the process and how to form a transformation and denoising cycle; Section IV presents our experimental results and analysis, demonstrating the improvement of our method on the Gaussian denoisers' ability to handle out-of-distribution noise; Section V concludes the paper, stating the limitations of our method and its future direction.

\section{Related work}
\subsection{Learning-based Denoising} 
Learning-based denoising methods can be classified into three categories: supervised, self-supervised, and zero-shot. Supervised denoising methods rely on training with pairs of noise and clean images. Early supervised denoising methods were based on CNN. DnCNN \cite{DnCNN} was the first method to apply CNN to Gaussian denoising. Subsequently, FFDNet \cite{FFDNet} achieved flexible denoising by adding an input of noise level map. Zhou et al. \cite{PD} proposed PD-denoising, which employed pixel-shuffle down-sampling to break the spatial correlation of real-world noise, enabling networks trained with Gaussian noise to remove real-world noise. DRUNet \cite{DRUNet} was a denoising network based on UNet. Zhang et al. \cite{SCUNet} combined DRUNet \cite{DRUNet} and SwinIR \cite{SwinIR} to propose SCUNet. They designed a noise synthesis method and trained a blind denoising model with the synthesized noise, which also has a good removal effect on real-world noise. Jiang et al. \cite{AdapPrior} introduced APDNet, which leveraged adaptive priors to improve the denoising performance and generalization of the network. Ding et al. \cite{Wavelet} proposed WACAFRN, combining adaptive coordinate attention mechanism and wavelet attention mechanism to capture and learn fine noise points more precisely and preserve edge information more effectively. In recent years, supervised denoising methods based on Transformer have been developed. Representative methods include Restormer \cite{Restormer}, UFormer \cite{Uformer}, etc. 

Self-supervised methods do not require clean images. A typical approach is blind spot networks, first proposed by Noise2Void \cite{N2V}. However, the original blind spot network is unsuitable for real-world noise removal due to noise's spatial correlation. Thus, AP-BSN \cite{AP-BSN} applied it to real-world scenarios via an asymmetric PD strategy. Li et al. \cite{TBSN} redesigned spatial and channel attention mechanisms and proposed a Transformer-based blind spot network (TBSN) to remove real-world noise. Yu et al. \cite{Inrapatch} proposed UBSN, which utilized randomized PD to effectively handle structural noise. Fan et al. \cite{ComBSN} introduced complementary blind-Spot network, which leverages controllable masked convolution to enable the network learn the ignored information of the central pixel. For zero-shot methods, ZS-N2N \cite{Zs-n2n} sampled two sub-noisy images from a single noisy image to train the denoising network. MASH \cite{MASH} proposed local pixel shuffling to break the spatial correlation of real-world noise. It was a zero-shot blind spot denoising method. Learning-based methods each have their own advantages, but generalization is a common problem. Our approach is based on learning-based methods, aiming to address their generalization issues.

\subsection{Generalization in Image Denoising}
Learning-based denoising methods often perform poorly when handling unseen noise types. For example, supervised models trained on Gaussian noise fail for real-world noise, while self-supervised models trained on real-world noise struggle with synthetic noises. Therefore, some studies have focused on improving the generalization of denoising networks. Mohan et al. \cite{BF} found that bias-free denoisers are more robust and can remove invisible noise levels even when trained within a limited noise range. MaskDenoising \cite{Maskde} proposed masked training to enhance generalization. Clipdenoising \cite{Clipde} used a pre-trained CLIP \cite{Clip} model to extract noise-robust deep features for denoising. LAN \cite{LAN} adjusted the noise distribution by adding noise to align it with the trained one, yet struggled to alter the noise's signal and spatial correlations.  DMID \cite{DMID} converted real-world noise to Gaussian noise via NN \cite{NN} and denoised using a pre-trained diffusion model; however, the NN has limited ability to convert many synthetic noises and requires extensive iterations. Ha et al. \cite{LTN} proposed a noise conversion network. It can effectively convert real-world noise into Gaussian noise, but it relies on supervised training and cannot convert other types of noise such as Poisson noise. Similar to the LTN \cite{LTN} method, our approach also attempts to transform the noise. The difference is that LTN \cite{LTN} is a learning-based method that constrains noise through loss functions in both the frequency domain and the spatial domain to make its distribution consistent with that of Gaussian noise. Our method is through histogram matching, a more direct approach. Although histogram matching is simple, experiments have proved that it is very effective.

\begin{figure*}[htbp]
\centering
\includegraphics[width=0.94\textwidth]{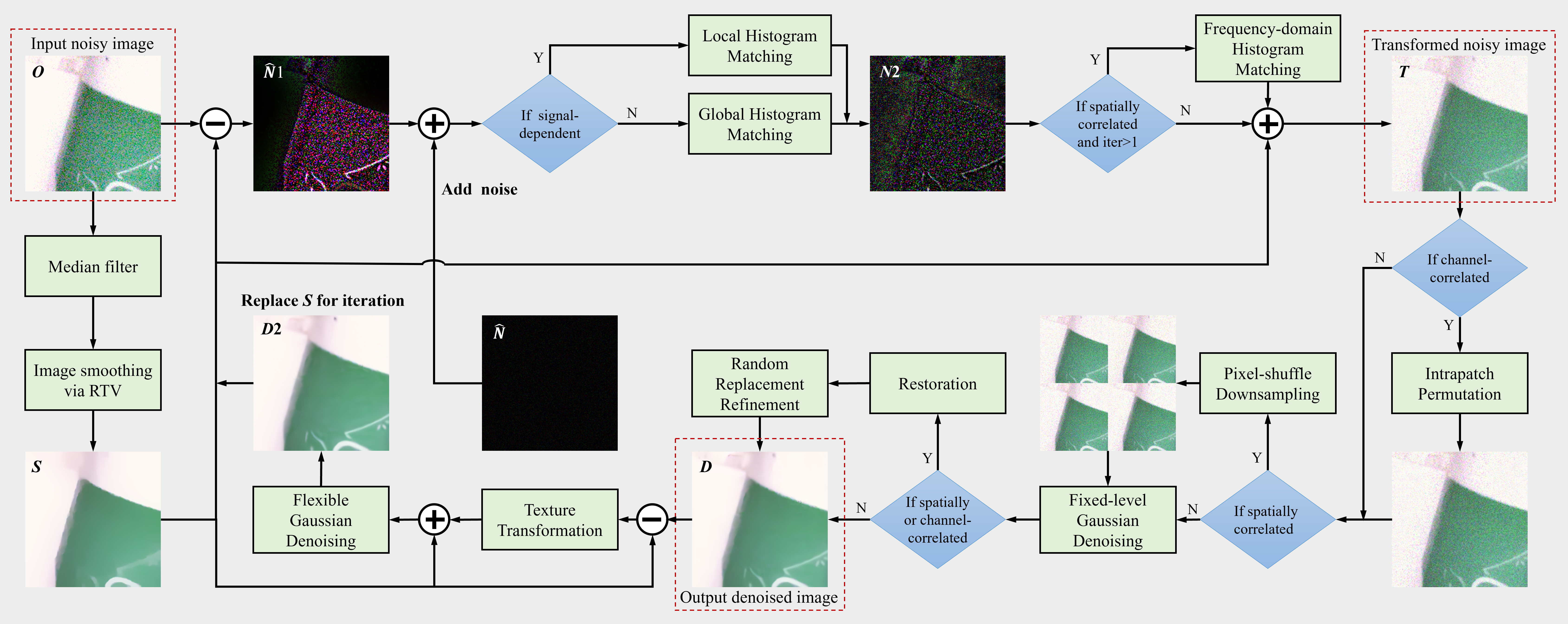}
\caption{Overall flowchart of our method. The initial noise for transformation is obtained by subtracting the smoothed image from the original noisy image. Subsequently, appropriate noise transformation and shuffling  strategies will be selected based on the signal, spatial and channel correlations of the noise. We then employ a fixed-level Gaussian denoiser to obtain the denoising result, which will be used for the next round of iteration after certain processing, forming a mutually reinforcing cycle.}
\label{fig2}
\end{figure*}

\section{Method}
The overall flowchart of our method is shown in Fig. \ref{fig2}. A noisy image is first smoothed using median filtering and RTV \cite{RTV}. Subtracting the smoothed image from the original noisy image gives the initial noise. This noise subsequently undergoes histogram-matching-based noise transformation, where local histogram matching is performed based on the noise's signal correlation, and frequency-domain histogram matching based on its spatial correlation. The transformed noise is then added to the smoothed image, resulting in the transformed noisy image. Based on the noise's channel and spatial correlations, a decision is made if to employ intrapatch permutation and/or PD. The transformed image is denoised via a fixed-level Gaussian denoiser. If intrapatch permutation or downsampling is applied, post-denoising restoration and refinement are necessary to recover the original order and eliminate artifacts. The final denoised result, following texture transformation and flexible denoising, replaces the original smoothed image for the next iteration, forming a mutually reinforcing cycle. Detailed illustrations of each step are provided below.

\subsection{Initial Noise Estimation}
To avoid noise residue, the initial noise used for noise transformation should contain all noise components, so we perform image smoothing operations. Prior to smoothing, there is a simple median filtering process, which is designed for non-zero mean noise like random impulsive noise. Subsequently, the RTV \cite{RTV} method is used for smoothing, which effectively separates image structure and texture to achieve smoothing. The formulas are given below:
\begin{equation}
\bm{O}1=Median\text{-}filter(\bm{O}),   
\end{equation}
\begin{equation}
\arg \underset{\bm{S}}{\mathop{\min }}\,\left\| \bm{S}-\bm{O}1 \right\|_{F}^{2}+{\alpha}\left\| {\bm{{W}}}\odot \nabla \bm{S} \right\|_{1},   
\end{equation}
where $\bm{S}$ is the smoothed map, ${{\left\|\cdot\right\|}_{\text{F}}}$ and ${{\left\|\cdot\right\|}_{\text{1}}}$ represent the Frobenius and $l_1$ norms, $\nabla$ represents the first-order differential operator, containing $\nabla_h$ (horizontal) and $\nabla_v$ (vertical), $\odot$ denotes element-wise multiplication of matrices, and $\bm{W}$ is pixel-wise weight. For each position, $\bm{W}$ is given in Eq. (3).

\begin{equation}
	\bm{{W}}(x)=\sum\limits_{y\in \Omega (X)}{\frac{{{G}_{\sigma }}(x,y)}{\left| \sum{_{y\in \Omega (X)}{{G}_{\sigma }}(x,y)\nabla {\bm{O}{1}}} \right|+\varepsilon }}.
\end{equation}

The ${G}_{\sigma }(x,y)$ is a Gaussian kernel with standard deviation $\sigma$. We can use an approximate approach to solve the Eq. (2), as shown in the following equations:
\begin{equation}
	\arg \underset{\bm{S}}{\mathop{\min }}\,\left\| \bm{S}-{\bm{O}{1}} \right\|_{F}^{2}+{\alpha }\sum\limits_{x}{{{\widehat{\bm{W}}}(x)}{{(\nabla {\bm{S}}(x))^{2}}}}.
\end{equation}
\begin{equation}
	{{\widehat{\bm{W}}}}(x)=\frac{{\bm{W}}(x)}{\left| \nabla {\bm{O}{1}(x)} \right|+\varepsilon }.
\end{equation}

Then, to solve Eq. (4), we can directly take the derivative. Define $\text{vec}(\cdot)$ as the matrix vectorization operator. Denote $\bm{s} = \text{vec}(\bm{S})$, $\bm{o}_1 = \text{vec}(\bm{O}1)$, and $\widehat{\bm{w}}= \text{vec}(\widehat{\bm{W}})$. Denote $\bm{D}$ be the Toeplitz matrices from the discrete gradient operators with forward difference. And we have $\bm{Ds}= \text{vec}(\nabla \bm{S})$. Further, we define the operator $Diag(\bm{x})$ to construct a diagonal matrix, whose main diagonal elements are vector $\bm{x}$. By differentiating Eq. (4) with respect to $\bm{S}$ and setting the derivative to 0, we can directly get the solution as follows:
\begin{equation}
	\bm{s}={(\bm{E}+{\alpha}{\bm{D}^{T}}Diag(\widehat{\bm{w}})\bm{D})}^{-1}\bm{o}_{1},
\end{equation}
where $\bm{E}\in {{\mathbb{R}}^{mn\times mn}}$ represents the identity matrix and $mn$ represents image resolution. Finally, $\bm{S}$ can be obtained by the inverse vectorization $\bm{S}=ve{{c}^{-1}}(\bm{s})$. To ensure a smooth image, the process is iterative. After obtaining $\bm{S}$, it will be used to calculate the new weight matrix $\bm{W}$. Subsequently, the new $\bm{S}$ will be obtained. Finally, the parameters are set as follows: $\alpha$ is 0.015, $\sigma$ is 3, and the iteration number is 4.

After obtaining the smoothed image $\bm{S}$, the initial noise $\widehat{\bm{N}}1$ is derived as follows:
\begin{equation}
\widehat{\bm{N}}1=\bm{O}-\bm{S}.
\end{equation}
\subsection{Noise Transformation}
Noise transformation is performed via histogram matching; however, unlike conventional image histogram matching, our noise histogram matching operates by interval and incorporates linear interpolation to achieve one-to-one mapping.
\subsubsection{Global histogram matching}
To make the noise transformation more robust, we first add a small amount of Gaussian noise, as follows:
\begin{equation}
{\bm{N}}1={\widehat{\bm{N}}}1+\widehat{\bm{N}},
\end{equation}
where $\widehat{\bm{N}}$ is the added Gaussian noise with a small standard deviation, taking $0.01$ default and ${\bm{N}}1$ is the noise to be transformed. We can assume that there is no noise in the image. In this case, noise must be added to make the transformation successful, otherwise, the transformation can only change the texture.

The target noise $\bm{N}0$ is Gaussian noise, as follows:
\begin{equation}
\bm{N}0\sim \mathcal{N}(0,{{\sigma}_{0}}),
\end{equation}
where ${\sigma}_{0}$ is the standard deviation.

The distribution of ${\bm{N}}1$ is generally different from $\bm{N}0$, but we can use the histogram matching method to make $\bm{N}1$ similar to $\bm{N}0$. First, we calculate the cumulative histogram distribution of $\bm{N}0$ and $\bm{N}1$:
\begin{equation}
\left\{ 
\begin{aligned}
  & (\bm{C}0,\bm{X}0)=\textbf{CDF}(\bm{N}0,B), \\ 
  & (\bm{C}1,\bm{X}1)=\textbf{CDF}(\bm{N}1,B), \\ 
\end{aligned} 
\right.
\end{equation}
where \textbf{CDF}($x$, $y$) denotes the cumulative distribution function, which computes the cumulative probability of $x$ based on specified interval division or interval count $y$. $\bm{X}1$ and $\bm{X}0$ represent the returned interval divisions. $\bm{C}1$ and $\bm{C}0$ are the returned cumulative distributions, corresponding to each interval of $\bm{X}1$ and $\bm{X}0$, respectively. $B$ represents the number of intervals.

Then, we add 0 to $\bm{C}0$ and $\bm{C}1$ to align them with $\bm{X}0$ and $\bm{X}1$:
\begin{equation}
\left\{ 
\begin{aligned}
  & \bm{C}0=[0,\bm{C}0], \\ 
  & \bm{C}1=[0,\bm{C}1]. \\
\end{aligned} 
\right.
\end{equation}

We then use linear interpolation to obtain the cumulative probability distribution for each pixel in $\bm{N}1$, as follows:
\begin{equation}
\bm{C}=\textbf{interp}(\bm{X}1,\bm{C}1,\bm{N}1),
\end{equation}
where $\textbf{interp}(x,y,z)$ is a linear interpolation function. And $x$ is the set of x-coordinates of known data points, $y$ is the set of y-coordinates of known data points, and $z$ is the coordinate point to interpolate. $\bm{C}$ is the cumulative probability corresponding to each pixel of $\bm{N}1$.

Finally, we can map $\bm{C}$ to $\bm{C}0$ to get the corresponding value after histogram matching, which is also obtained by linear interpolation, as follows:
\begin{equation}
\bm{N}2=\textbf{interp}(\bm{C}0,\bm{X}0,\bm{C}).
\end{equation}

The transformed noise $\bm{N}2$ is obtained. $\bm{N}2$ is then added to the $\bm{S}$ to get the transformed noisy image $\bm{T}$ , as follows:
\begin{equation}
\bm{T}=\bm{S}+\bm{N}2.
\end{equation}

\subsubsection{Local histogram matching}
To handle signal-dependent or local noise, we use local histogram matching. We just divide the noise map $\bm{N}1$ into blocks of size $b \times b$ with $k$ pixels overlapping. We then perform histogram matching on each block. The process is consistent with global histogram matching, except that local histogram matching uses the specified interval division to calculate the cumulative distribution.

\subsubsection{Frequency-domain histogram matching}
To handle spatially correlated noise, we extend spatial-domain histogram matching to frequency domain. Let $\bm{N}0_f$ and $\bm{N}2_f$ be the Fourier transforms of $\bm{N}0$ and $\bm{N}2$. We then calculate the cumulative distribution in the frequency domain:
\begin{equation}
\left\{ 
\begin{aligned}
  & (\bm{C}0_f,\bm{X}0_f)=\textbf{CDF}(\textbf{real}(\bm{N}0_f),B), \\ 
  & (\bm{C}2_r,\bm{X}2_r)=\textbf{CDF}(\textbf{real}(\bm{N}2_f),B), \\ 
  & (\bm{C}2_j,\bm{X}2_j)=\textbf{CDF}(\textbf{imag}(\bm{N}2_f),B), \\ 
\end{aligned} 
\right.
\end{equation}
where \textbf{real}() and \textbf{imag}() extract the real and imaginary parts; $\bm{C}0_f$ and $\bm{X}0_f$ are the cumulative distribution and intervals of the real part of $\bm{N}0_f$; $\bm{C}2_r$ and $\bm{X}2_r$ are those of the real part of $\bm{N}2_f$; and $\bm{C}2_j$ and $\bm{X}2_j$ are those of the imaginary part of $\bm{N}2_f$. Given that the imaginary and real parts of $\bm{N}0_f$ follow the same Gaussian distribution, we only need to consider the distribution of the real part. Similarly, we have:
\begin{equation}
\left\{ 
\begin{aligned}
  & \bm{C}0_f=[0,\bm{C}0_f], \\ 
  & \bm{C}2_r=[0,\bm{C}2_r], \\
  & \bm{C}2_j=[0,\bm{C}2_j].
\end{aligned} 
\right.
\end{equation}

Then, we perform frequency-domain histogram matching:
\begin{equation}
\left\{ 
\begin{aligned}
&\bm{C}_r=\textbf{interp}(\bm{X}2_r,\bm{C}2_r,\textbf{real}(\bm{N}2_f)),\\
&\bm{C}_j=\textbf{interp}(\bm{X}2_j,\bm{C}2_j,\textbf{imag}(\bm{N}2_f)),\\
&\bm{N}2_r=\textbf{interp}(\bm{C}0_f,\bm{X}0_f,\bm{C}_r),\\
&\bm{N}2_j=\textbf{interp}(\bm{C}0_f,\bm{X}0_f,\bm{C}_j),
\end{aligned} 
\right.
\end{equation}
where $\bm{N}2_r$ and $\bm{N}2_j$ are the real and imaginary parts of $\bm{N}2$ after frequency-domain histogram matching. It is acquired by the inverse Fast Fourier Transform (IFFT):
\begin{equation}
\bm{N}2=IFFT(\bm{N}{{2}_{r}}+j\bm{N}{{2}_{j}}).
\end{equation}

Frequency-domain histogram matching can reduce the spatial correlation of noise but may also damage textures. Since the initial noise contains many textures, we only determine whether to perform frequency-domain histogram matching when the iteration is greater than 1.

\subsection{Intrapatch Permutation}
Intrapatch permutation is proposed in the paper \cite{Inrapatch} to address spatially-correlated noise. It divides the image into $m\times m$ patches and then randomly shuffles each patch. The shuffled order is recorded for subsequent restoration. Our intrapatch permutation focuses on addressing channel correlation, which is performed across three channels. In our approach, the shuffling order of the three channels within each patch is random, whereas in the study \cite{Inrapatch}, it remains consistent.

\subsection{Pixel-shuffle Down-sampling}
For spatially correlated noise, we adopt pixel-shuffle down-sampling \cite{PD} to further break the spatial correlation on the basis of frequency-domain histogram matching. It downsamples the noisy image into four subimages and then assembles them into one image, thus reducing spatial correlation. When noise exhibits strong spatial correlation, the noisy image after PD may still retain such correlation. Therefore, we have frequency-domain histogram matching before PD. Their combination can handle spatially correlated noise well.

\subsection{Fixed-level Gaussian Denoising}
While the transformed noise may still deviate from Gaussian noise, its noise level is constrained within a specific range. Therefore, we use fixed-level Gaussian denoising. It can be a denoiser trained with fixed-level Gaussian noise or a flexible denoiser given a specified noise level. And the denoising level is consistent with the target Gaussian noise level. The formula is as follows:
\begin{equation}
    \bm{D}=Fixed\text{-}denoiser(\bm{T},{{\sigma }_{0}}),
\end{equation}
where $\bm{D}$ is the denoised image, and $\bm{T}$ is the transformed noisy image, which may have undergone intrapatch permutation or pixel-shuffle down-sampling.

\subsection{Restoration and Refinement}
Intrapatch permutation and pixel-shuffle down-sampling disrupt the order of pixels, so restoration is required after denoising. The restored image may have artifacts and needs to be refined. We follow \cite{AP-BSN}
to use Random Replacement Refinement. In our method, the restored image will be randomly replaced by the transformed noisy image with a probability of $p$. The denoiser used for the refinement is still a fixed-level Gaussian denoiser.

\subsection{Texture Transformation}
The purpose of texture transformation is to convert back the texture that has been changed during the noise transformation process. Let $\bm{t}1$ and $\bm{t}2$ be the textures contained in the noises $\bm{N}1$ and $\bm{N}2$. Since $\bm{N}1$ to $\bm{N}2$ is a pixel-to-pixel mapping and we consider that the signal-to-noise ratio of $\bm{N}1$ and $\bm{N}2$ is consistent before and after the transformation, we have:
\begin{equation}
\bm{t}1/\bm{N}1=\bm{t}2/\bm{N}2.
\end{equation}

Then, the formulas for texture transformation are as follows:
\begin{equation}
\left\{
\begin{aligned}
 & \bm{t}2=\bm{D}-\bm{S},\\
 & \bm{R}=\bm{t}2/\bm{N}2,\\
 & \bm{t}1=\bm{N}1\odot \bm{R}, \\ 
 & \bm{t}1(\bm{R}<-1\vee \bm{R}>1)=t2(\bm{R}<-1\vee \bm{R}>1),
\end{aligned}
\right.
\end{equation}
where $\bm{R}$ can be regarded as the signal-to-noise ratio of each pixel in $\bm{N}2$.

The transformed back texture $\bm{t}1$ is then added to $\bm{S}$ to get the texture transformation result $\bm{D}1$:
\begin{equation}
\bm{D}1=\bm{S}+\bm{t}1.
\end{equation}

\subsection{Flexible Gaussian Denoising}
The texture transformation result $\bm{D}1$ may still contain noise due to potential inadequacies in the prior denoising step. Additional denoising is therefore required. Here, we use flexible Gaussian denoising. Due to the uncertainty of the noise level, we roughly regard $\bm{t}1$ as a noise level map and input it into the flexible denoiser as follows:
\begin{equation}
    \bm{D}2=Flex\text{-}denoiser(\bm{D}1,\textbf{abs}(\bm{t}1)),
\end{equation}
where \textbf{abs}() is the absolute value function and $\bm{D}2$ is the denoising result of $\bm{D}1$. Finally, $\bm{D}2$ replaces $\bm{S}$ for the next round of noise transformation, forming a cycle.

\begin{figure*}[htbp]
	\centering
	\includegraphics[width=0.98\textwidth]{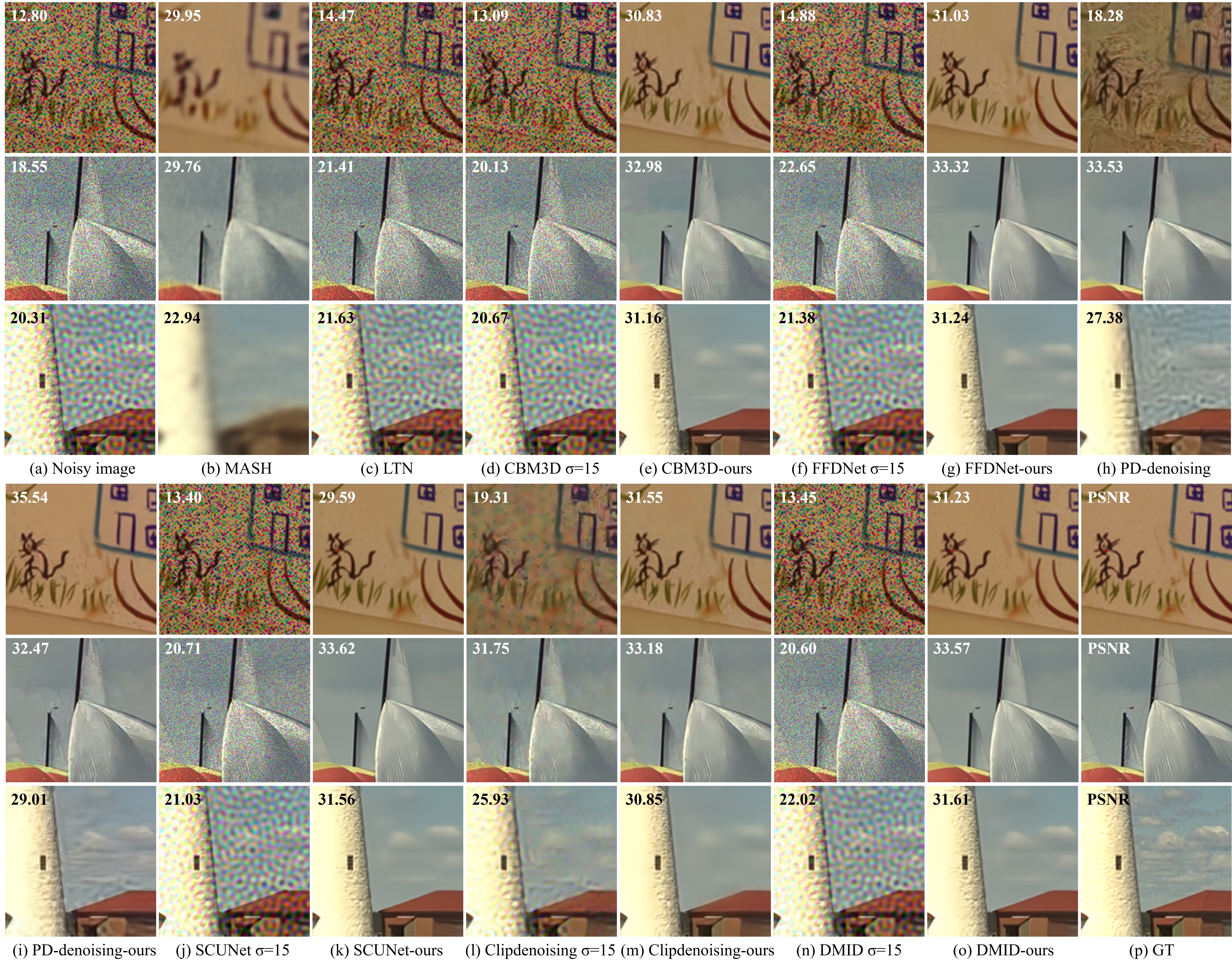}
	\caption{Qualitative comparison on synthetic noise. The noise images from the first row to the last row are respectively Bernoulli noise, speckle noise and circular repeating pattern noise images. They are all out-of-distribution noises that Gaussian denoisers are difficult to remove.}
	\label{fig3}
\end{figure*}

\begin{figure*}[htbp]
	\centering
	\includegraphics[width=0.9\textwidth]{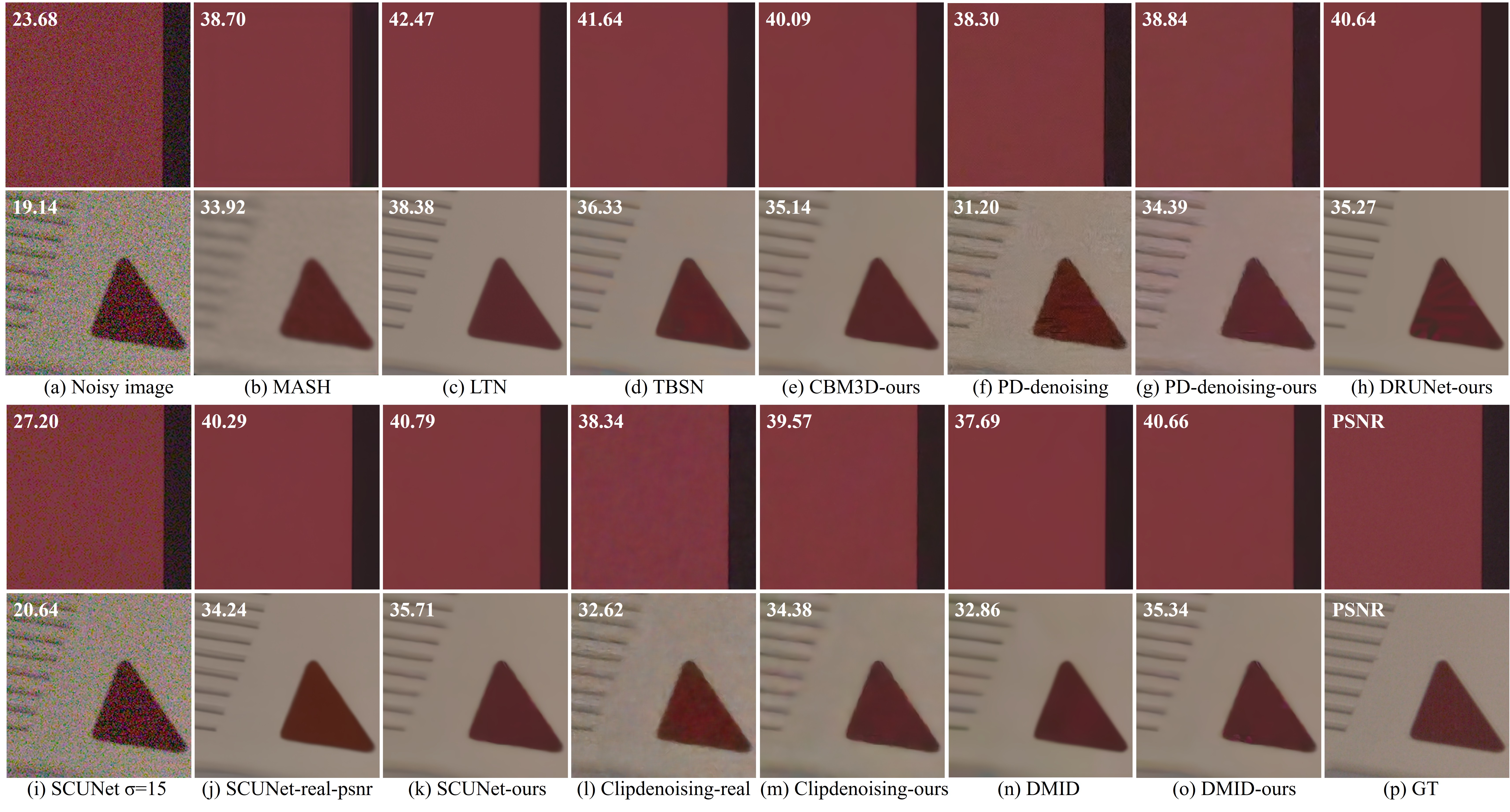}
	\caption{Qualitative comparison on real-world noise. Real-world noise is much more complex than synthetic noise. Gaussian denoisers generally have difficulty removing real-world noise. But through our noise transformation, they have achieved excellent denoising results.}
	\label{fig4}
\end{figure*}

\section{Experiments}
\subsection{Implementation Details}
The parameters of our method are summarized as follows: \{$\sigma_0,B,b,k,m,p$\} = \{$15/255, 2000, 36, 4, 2, 0.3$\}, the number of Random Replacement Refinement is 4, and the number of iterations is 3. Moreover, the flexible denoiser of our method is FFDNet \cite{FFDNet} by default. To verify the effectiveness of the method, we use synthetic noise and real-world noise. For synthetic noise, we synthesized various out-of-distribution noise on the Kodak24 \cite{Kodak} and McMaster \cite{McMater} datasets, including global noise: Gaussian noise with $\sigma=25$, uniform noise with $x=0.3$, salt-and-pepper noise with $d=0.2$, random impulse noise with $d=0.2$, and Bernoulli noise with $d=0.2$; Signal-dependent noise: Poisson noise with $\lambda=25$ and speckle noise with $\sigma=55$, as well as spatially correlated noise: circular repeating pattern noise with $\sigma=25$, which is a kind of spatial Gaussian noise with circular repetitive pattern. Uniform noise with $x=0.3$ follows a uniform distribution of [-0.3,0.3]. Speckle noise with $\sigma=55$ is obtained by multiplying the clean image by Gaussian noise with $\sigma=55$. For real-world noise, we use the SIDD-validation dataset \cite{SIDD}. 

We adopt an appropriate strategy based on the noise properties. Specifically, global histogram matching is used for global noise; local histogram matching for signal-dependent noise; and frequency-domain histogram matching combined with PD for spatially dependent noise. For real-world noise, since it simultaneously exhibits signal, spatial, and channel correlations, we integrate local histogram matching, frequency-domain histogram matching with PD, and intrapatch permutation. However, the SIDD-validation dataset contains low-light noisy images, where signal-independent noise constitutes the main component \cite{Noise_model}. Accordingly, local histogram matching is not required here. We set the brightness threshold to 0.2, such that noisy images with a mean brightness below this threshold undergo global histogram matching.

Regarding the comparison method, for PD-denoising \cite{PD}, AP-BSN \cite{AP-BSN} and TBSN \cite{TBSN},  we do not use pixel-shuffle down-sampling and refinement for spatially independent noise, as they are not applicable to such noise. And PD-denoising is coupled with its noise level estimator for denoising, while PD-denoising $\sigma=15$ does not perform noise estimation and uses a fixed noise level. The MASH \cite{MASH} method has an adaptive masking strategy and a local shuffling strategy designed for spatially correlated noise. However, these two strategies are not suitable for spatially independent noise. Consequently, for spatially independent noise, we utilize the baseline of MASH with a masking rate of 0.1. And the adaptive masking and local shuffling strategies are employed for spatially correlated noise. For the DMID \cite{DMID} method, we first use its noise transformation method to convert the noise, and then denoise the converted noisy image with a diffusion model. DMID $\sigma=15$ does not perform noise conversion and uses a diffusion model with fixed timesteps for denoising. Finally, we all use the pre-trained models and default parameters provided by the authors.

\subsection{Results}
\subsubsection{Qualitative comparison}
Figs. \ref{fig3} and \ref{fig4} present the qualitative comparison results. In Fig. \ref{fig3}, input noisy images are respectively Bernoulli noise, speckle noise and circular repeating pattern noise images which are all challenging for Gaussian denoisers to remove. However, with our method, Gaussian denoisers achieve excellent denoising performance. Fig. \ref{fig4} shows the denoising effect of real-world noise which is more complex. Our method still removes noise well.

\subsubsection{Quantitative comparison}

\begin{table*}[htbp]
     \caption{Quantitative comparison on the Kodak24 and SIDD-validation datasets. Our method employs a Gaussian denoiser with $\sigma=15$. (PD-de and Clip-de are abbreviations for PD-denoising and Clipdenoising respectively.)}
    \label{table1}
    \centering
    \setlength{\tabcolsep}{0.9mm}
    
    \small
    \begin{tabular}{cccccccccc}
    \toprule
        Noise Types& \makecell[c]{Gaussian\\$\sigma=25$}& \makecell[c]{Uniform\\$x=0.3$}&\makecell[c]{ S\&P\\$d=0.2$} & \makecell[c]{Impulse\\$d=0.2$} & \makecell[c]{Bernoulli\\ $d=0.2$} &  \makecell[c]{Poisson\\$\lambda=25$} &\makecell[c]{Speckle\\ $\sigma=55$} & \makecell[c]{Circular\\$\sigma=25$} & \makecell[c]{Real-wold\\noise}
       \\ \hline \hline
        AP-BSN & 27.65/0.750 & 25.08/0.646 & 22.12/0.478 & 22.86/ 0.562 & 21.44/0.685 & 26.36/0.706 & 27.33/0.760 & 25.91/\textbf{0.690} & 36.74/0.888  \\
        MASH & \textbf{28.02}/\textbf{0.799} & \textbf{26.55}/\textbf{0.728} & \textbf{25.66}/\textbf{0.712} & \textbf{26.15}/\textbf{0.740} & \textbf{27.52}/\textbf{0.813} & \textbf{27.60}/\textbf{0.780} & \textbf{28.29}/\textbf{0.816} & 23.50/0.606 & 35.06/0.851  \\ 
        LTN & 24.29/0.493 & 20.77/0.364 & 11.38/0.092 & 17.41/0.253 & 15.18/0.305 & 20.69/0.392 & 23.63/0.570 & 21.81/0.410 & \textbf{39.24}/\textbf{0.916}  \\ 
        TBSN & 27.68/0.754 & 24.81/0.649 & 20.00/0.430 & 22.11/0.566 & 21.58/0.634 & 26.14/0.691 & 27.19/0.732 & \textbf{25.94}/0.677 & 37.70/0.896  \\ 
        \hline \hline
        CBM3D $\sigma$=15 & 23.92/0.476 & 16.09/0.190 & 12.33/0.116 & 16.31/0.230 & 14.11/0.285 & 19.59/0.370 & 23.12/0.607 & 20.82/0.366 & 29.04/0.596  \\ 
        CBM3D-ours & \textbf{31.81}/\textbf{0.870} & \textbf{29.78}/\textbf{0.825} & \textbf{32.38}/\textbf{0.910} & \textbf{32.28}/\textbf{0.914} & \textbf{27.10}/\textbf{0.886} & \textbf{30.04}/\textbf{0.843} & \textbf{31.42}/\textbf{0.881} & \textbf{31.60}/\textbf{0.860} & \textbf{35.95}/\textbf{0.897}  \\ 
        \hline \hline
        DnCNN $\sigma$=25 & \textbf{32.24}/\textbf{0.880} & 19.56/0.300 & 13.93/0.152 & 18.75/0.359 & 15.50/0.405 & 24.95/0.616 & 28.00/0.755 & 23.33/0.480 & 31.97/0.711  \\ 
        DnCNN $\sigma$=15 & 24.58/0.513 & 16.52/0.205 & 12.77/0.125 & 16.62/0.248 & 14.36/0.312 & 20.23/0.404 & 23.56/0.613 & 21.70/0.404 & 29.06/0.600  \\ 
        DnCNN-ours & 32.12/0.877 & \textbf{30.17}/\textbf{0.836} & \textbf{31.85}/\textbf{0.908} & \textbf{32.13}/\textbf{0.913} & \textbf{26.91}/\textbf{0.879} & \textbf{30.19}/\textbf{0.849} & \textbf{31.51}/\textbf{0.883} & \textbf{31.62}/\textbf{0.867} & \textbf{35.59}/\textbf{0.895}  \\ 
        \hline \hline
        FFDNet $\sigma$=25 & \textbf{32.13}/\textbf{0.878} & 20.86/0.358 & 15.80/0.191 & 20.91/0.442 & 16.88/0.446 & 25.98/0.657 & 28.57/0.770 & 23.21/0.470 & 32.28/0.717  \\ 
        FFDNet $\sigma$=15 & 25.93/0.584 & 18.72/0.279 & 14.92/0.167 & 18.68/0.306 & 15.99/0.361 & 22.00/0.467 & 25.11/0.653 & 21.57/0.399 & 29.22/0.606  \\ 
        FFDNet-ours & 32.01/0.874 & \textbf{30.12}/\textbf{0.834} & \textbf{32.68}/\textbf{0.914} & \textbf{32.64}/\textbf{0.918} & \textbf{27.91}/\textbf{0.896} & \textbf{30.19}/\textbf{0.849} & \textbf{31.56}/\textbf{0.888} & \textbf{31.56}/\textbf{0.859} & \textbf{35.88}/\textbf{0.897}  \\ 
        \hline \hline
        DRUNet $\sigma$=25  & \textbf{32.79}/\textbf{0.892} & 18.84/0.268 & 13.66/0.139 & 19.00/0.342 & 15.44/0.383 & 24.05/0.580 & 27.27/0.741 & 22.13/0.391 & 31.81/0.695  \\ 
        DRUNet $\sigma$=15  & 24.12/0.484 & 16.69/0.208 & 12.74/0.125 & 16.86/0.248 & 14.43/0.300 & 20.15/0.390 & 23.36/0.601 & 21.30/0.377 & 28.10/0.573  \\ 
        DRUNet-ours & 32.57/0.887 & \textbf{30.59}/\textbf{0.849} & \textbf{32.50}/\textbf{0.914} & \textbf{32.53}/\textbf{0.918} & \textbf{26.57}/\textbf{0.874} & \textbf{30.45}/\textbf{0.861} & \textbf{31.70}/\textbf{0.892} & \textbf{31.97}/\textbf{0.873} & \textbf{35.84}/\textbf{0.897}  \\ 
        \hline \hline
        PD-de & \textbf{31.56}/0.863 & \textbf{28.74}/\textbf{0.788} & 27.49/0.788 & \textbf{40.20}/\textbf{0.982} & 18.95/0.624 & \textbf{30.37}/\textbf{0.843} & \textbf{31.80}/\textbf{0.887} & 27.82/0.720 & 33.97/0.830  \\ 
        PD-de $\sigma$=15 & 30.07/0.788 & 23.41/0.467 & 25.46/0.665 & 30.66/0.844 & 22.00/0.598 & 26.88/0.662 & 29.34/0.785 & 26.71/0.641 & 32.69/0.715  \\ 
        PD-de-ours & 31.15/\textbf{0.868} & 27.74/0.784 & \textbf{32.60}/\textbf{0.920} & 33.02/0.929 & \textbf{32.41}/\textbf{0.933} & 29.55/0.839 & 30.93/0.882 & \textbf{29.03}/\textbf{0.825} & \textbf{34.90}/\textbf{0.867}  \\ 
        \hline \hline
        Clip-de-real & 24.79/0.741 & 18.89/0.528 & 16.61/0.374 & 20.55/0.566 & 20.58/0.658 & 20.74/0.613 & 24.45/0.742 & 25.56/0.583 & 34.82/0.867  \\ 
        Clip-de $\sigma$=15 & 31.77/0.870 & 28.27/0.763 & 22.57/0.570 & 24.00/0.695 & 19.43/0.666 & 29.61/0.824 & 30.64/0.860 & 26.92/0.713 & 30.21/0.641  \\ 
        Clip-de-ours & \textbf{32.05}/\textbf{0.878} & \textbf{30.05}/\textbf{0.835} & \textbf{32.53}/\textbf{0.915} & \textbf{32.56}/\textbf{0.920} & \textbf{28.50}/\textbf{0.914} & \textbf{30.08}/\textbf{0.852} & \textbf{31.40}/\textbf{0.887} & \textbf{30.99}/\textbf{0.851} & \textbf{35.20}/\textbf{0.893}  \\ 
        \hline \hline
        Restormer-real & 26.45/0.764 & 20.57/0.625 & 16.90/0.346 & 20.97/0.530 & \textbf{25.26}/0.711 & 23.08/0.632 & 24.79/0.704 & 25.30/0.577 & \textbf{40.02}/\textbf{0.922}  \\ 
        Restormer $\sigma$=25  & \textbf{32.94}/\textbf{0.894} & 22.42/0.405 & 16.09/0.187 & 19.02/0.360 & 16.02/0.405 & 26.28/0.671 & 27.40/0.756 & 20.31/0.359 & 24.10/0.422  \\ 
        Restormer $\sigma$=15  & 27.58/0.645 & 21.30/0.378 & 17.50/0.238 & 20.28/0.431 & 16.13/0.401 & 23.10/0.543 & 24.72/0.660 & 20.00/0.360 & 22.54/0.370  \\ 
        Restormer-ours  & 32.80/0.893 & \textbf{30.76}/\textbf{0.854} & \textbf{30.56}/\textbf{0.893} & \textbf{30.98}/\textbf{0.899} & 24.57/\textbf{0.832} & \textbf{30.06}/\textbf{0.861} & \textbf{31.14}/\textbf{0.889} & \textbf{30.14}/\textbf{0.858} & 34.84/0.867  \\ 
        \hline \hline
        SCUNet-real & 31.38/0.868 & 28.35/0.794 & 21.56/0.475 & 23.76/0.586 & 22.84/0.756 & 29.70/0.856 & 30.72/0.888 & 28.10/0.750 & 35.14/0.870  \\ 
        SCUNet $\sigma$=25  & \textbf{32.94}/\textbf{0.894} & 19.20/0.282 & 13.99/0.143 & 18.33/0.314 & 15.42/0.369 & 24.44/0.594 & 27.65/0.747 & 22.11/0.395 & 31.24/0.682  \\ 
        SCUNet $\sigma$=15  & 24.23/0.491 & 16.61/0.205 & 12.86/0.125 & 16.48/0.232 & 14.34/0.287 & 20.19/0.391 & 23.36/0.600 & 21.14/0.371 & 27.63/0.560  \\ 
        SCUNet-ours  & 32.74/0.889 & \textbf{30.74}/\textbf{0.851} & \textbf{31.55}/\textbf{0.905} & \textbf{31.98}/\textbf{0.913} & \textbf{26.16}/\textbf{0.866} & \textbf{30.54}/\textbf{0.863} & \textbf{31.70}/\textbf{0.893} & \textbf{32.01}/\textbf{0.872} & \textbf{36.04}/\textbf{0.899}  \\ 
        \hline \hline
        DMID  & 29.21/0.769 & 27.90/0.733 & 15.05/0.167 & 17.76/0.315 & 14.61/0.323 & 25.95/0.640 & 27.32/0.709 & 25.32/0.542 & 33.41/ \textbf{0.913}  \\ 
        DMID  $\sigma$=15 & 23.97/0.474 & 16.83/0.210 & 12.85/0.125 & 16.40/0.231 & 14.34/0.296 & 20.16/0.388 & 23.06/0.582 & 22.13/0.397 & 27.55/0.531  \\ 
        DMID-ours & \textbf{32.83}/\textbf{0.891} & \textbf{30.84}/\textbf{0.855} & \textbf{30.74}/\textbf{0.886} & \textbf{31.58}/\textbf{0.900} & \textbf{27.03}/\textbf{0.828} & \textbf{30.49}/\textbf{0.863} & \textbf{31.50}/\textbf{0.888} & \textbf{32.13}/\textbf{0.877} & \textbf{36.20}/0.899  \\
        \bottomrule
    \end{tabular}
\end{table*}

\begin{table*}[htbp]
     \caption{Quantitative comparison on the McMaster dataset. Our method employs a Gaussian denoiser with $\sigma=15$.}
    \label{table2}
    \centering
    \renewcommand\arraystretch{1.2}
    \setlength{\tabcolsep}{1mm}
    \small
    \begin{tabular}{ccccccccc}
    \toprule
        Noise Types& \makecell[c]{Gaussian\\$\sigma=25$}& \makecell[c]{Uniform\\$x=0.3$}&\makecell[c]{ S\&P\\$d=0.2$} & \makecell[c]{Impulse\\$d=0.2$} & \makecell[c]{Bernoulli\\ $d=0.2$} &  \makecell[c]{Poisson\\$\lambda=25$} &\makecell[c]{Speckle\\ $\sigma=55$} & \makecell[c]{Circular\\$\sigma=25$} 
       \\ \hline \hline
        AP-BSN & 28.15/0.755 & 25.25/0.641 & 21.48/0.481 & 22.46/0.556 & 21.83/0.702 & 27.12/0.728 & 28.18/0.777 & 25.80/\textbf{0.687}  \\
        MASH & \textbf{28.82}/\textbf{0.826} & \textbf{27.19}/\textbf{0.762} & \textbf{26.45}/\textbf{0.773} & \textbf{26.92}/\textbf{0.788} & \textbf{28.70}/\textbf{0.853} & \textbf{28.40}/\textbf{0.783} & \textbf{29.18}/\textbf{0.849} & 23.72/0.660  \\ 
        LTN & 24.70/0.513 & 19.08/0.285 & 12.05/0.113 & 15.96/0.225 & 15.72/0.401 & 21.66/0.507 & 24.08/0.639 & 21.85/0.407  \\ 
        TBSN & 27.68/0.724 & 24.11/0.594 & 18.27/0.386 & 22.24/0.504 & 22.01/0.632 & 26.68/0.697 & 27.84/0.740 & \textbf{25.88}/0.662  \\ 
        \hline \hline
        CBM3D $\sigma=15$ & 24.14/0.478 & 16.59/0.194 & 11.74/0.112 & 14.94/0.201 & 14.91/0.439 & 20.86/0.520 & 23.59/0.664 & 21.23/0.374  \\ 
        CBM3D-ours & \textbf{31.76}/\textbf{0.871} & \textbf{29.84}/\textbf{0.827} & \textbf{32.44}/\textbf{0.894} & \textbf{31.62}/\textbf{0.879} & \textbf{28.25}/\textbf{0.838} & \textbf{30.18}/\textbf{0.857} & \textbf{31.46}/\textbf{0.883} & \textbf{30.95}/\textbf{0.855}  \\ 
        \hline \hline
        DnCNN $\sigma=25$ & \textbf{32.48}/\textbf{0.889} & 20.12/0.314 & 13.17/0.143 & 16.81/0.289 & 15.88/0.509 & 25.58/0.682 & 28.03/0.766 & 24.10/0.531  \\ 
        DnCNN $\sigma=15$ & 24.90/0.525 & 17.12/0.213 & 12.16/0.120 & 15.22/0.215 & 15.04/0.462 & 21.33/0.548 & 23.94/0.680 & 22.17/0.421  \\ 
        DnCNN-ours & 32.29/0.884 & \textbf{30.46}/\textbf{0.850} &\textbf{32.32}/\textbf{0.896} & \textbf{31.81}/\textbf{0.881} & \textbf{27.88}/\textbf{0.828} & \textbf{30.34}/\textbf{0.859} & \textbf{31.62}/\textbf{0.886} & \textbf{31.19}/\textbf{0.860}  \\ 
        \hline \hline
        FFDNet $\sigma=25$ & \textbf{32.35}/\textbf{0.886} & 21.59/0.410 & 15.12/0.176 & 18.92/0.350 & 17.42/0.538 & 26.65/0.711 & 28.90/0.785 & 23.91/0.521  \\ 
        FFDNet $\sigma=15$ & 26.43/0.610 & 19.42/0.308 & 14.34/0.156 & 17.30/0.262 & 16.64/0.488 & 23.13/0.593 & 25.58/0.705 & 22.23/0.433  \\ 
        FFDNet-ours & 32.17/0.880 & \textbf{30.39}/\textbf{0.844} & \textbf{32.95}/\textbf{0.901} & \textbf{32.25}/\textbf{0.887} & \textbf{29.25}/\textbf{0.853} & \textbf{30.42}/\textbf{0.860} & \textbf{31.83}/\textbf{0.892} & \textbf{30.86}/\textbf{0.846}  \\
        \hline \hline
        DRUNet $\sigma=25$  & \textbf{33.16}/\textbf{0.904} & 19.67/0.317 & 12.86/0.132 & 16.97/0.287 & 16.08/0.504 & 24.77/0.664 & 27.30/0.756 & 22.55/0.425  \\ 
        DRUNet $\sigma=15$  & 24.62/0.515 & 17.56/0.233 & 12.09/0.119 & 15.40/0.217 & 15.19/0.455 & 21.29/0.544 & 23.85/0.676 & 21.77/0.394  \\ 
        DRUNet-ours & 32.95/0.899 & \textbf{30.99}/\textbf{0.863} & \textbf{32.99}/\textbf{0.906} & \textbf{32.26}/\textbf{0.891} & \textbf{27.49}/\textbf{0.825} & \textbf{30.85}/\textbf{0.873} & \textbf{32.12}/\textbf{0.897} & \textbf{31.80}/\textbf{0.869}  \\ 
        \hline \hline
        PD-denoising & \textbf{30.93}/\textbf{0.848} & \textbf{28.38}/\textbf{0.784} & 28.09/0.805 & \textbf{36.62}/\textbf{0.958} & 19.88/0.677 & \textbf{30.00}/0.812 & \textbf{31.50}/\textbf{0.896} & 27.56/0.730  \\ 
        PD-denoising $\sigma=15$ & 29.95/0.787 & 23.45/0.462 & 26.78/0.727 & 29.84/0.821 & 23.90/0.711 & 27.44/0.690 & 29.59/0.801 & 26.56/0.645  \\ 
        PD-denoising-ours & 30.51/0.838 & 26.34/0.716 & \textbf{32.41}/\textbf{0.903} & 32.26/0.897 & \textbf{32.25}/\textbf{0.906} & 29.43/\textbf{0.833} & 30.79/0.874 & \textbf{28.01}/\textbf{0.768}  \\ 
        \hline \hline
        Clipdenoising-real & 25.01/0.657 & 20.82/0.505 & 18.10/0.381 & 21.00/0.505 & 20.44/0.590 & 22.36/0.581 & 24.96/0.711 & 25.22/0.569  \\ 
        Clipdenoising $\sigma=15$ & 31.50/0.866 & 28.31/0.766 & 21.23/0.511 & 22.30/0.612 & 19.94/0.677 & 29.61/0.830 & 30.29/0.832 & 27.34/0.728  \\ 
        Clipdenoising-ours & \textbf{31.74}/\textbf{0.874} & \textbf{29.92}/\textbf{0.834} & \textbf{32.42}/\textbf{0.894} & \textbf{31.69}/\textbf{0.879} & \textbf{29.92}/\textbf{0.890} & \textbf{30.15}/\textbf{0.860} & \textbf{31.43}/\textbf{0.886} & \textbf{29.48}/\textbf{0.813}  \\ 
        \hline \hline
        Restormer-real & 26.89/0.765 & 21.15/0.592 & 16.06/0.315 & 19.18/0.480 & 25.32/0.718 & 23.15/0.620 & 24.60/0.681 & 25.44/0.593  \\ 
        Restormer $\sigma=25$  & \textbf{33.36}/\textbf{0.907} & 22.63/0.428 & 15.05/0.171 & 17.13/0.302 & 16.76/0.514 & 25.44/0.693 & 26.29/0.748 & 20.90/0.367  \\ 
        Restormer $\sigma=15$  & 27.34/0.635 & 21.23/0.381 & 16.12/0.211 & 18.88/0.367 & 16.20/0.493 & 22.77/0.606 & 24.41/0.697 & 20.87/0.373  \\ 
        Restormer-blind  & 33.33/0.906 & 30.85/0.864 & 21.08/0.566 & 21.29/0.594 & 17.48/0.595 & 26.75/0.753 & 25.88/0.757 & 20.89/0.376  \\ 
        Restormer-ours  & 33.20/0.903 & \textbf{31.15}/\textbf{0.867} & \textbf{30.35}/\textbf{0.877} & \textbf{29.63}/\textbf{0.852} & \textbf{25.38}/\textbf{0.753} & \textbf{30.44}/\textbf{0.847} & \textbf{31.23}/\textbf{0.864} & \textbf{29.33}/\textbf{0.842}  \\ 
        \hline \hline
        SCUNet-real-psnr & 30.78/0.786 & 27.78/0.695 & 24.30/0.600 & 24.89/0.639 & 23.06/0.692 & 30.18/0.804 & 31.46/0.862 & 27.50/0.686  \\ 
        SCUNet $\sigma=25$  & \textbf{33.36}/\textbf{0.907} & 19.87/0.307 & 13.28/0.137 & 16.59/0.269 & 16.06/0.493 & 24.98/0.669 & 27.37/0.759 & 22.51/0.421  \\ 
        SCUNet $\sigma=15$  & 24.57/0.505 & 17.32/0.218 & 12.31/0.121 & 15.16/0.207 & 15.11/0.447 & 21.22/0.541 & 23.72/0.673 & 21.52/0.383  \\ 
        SCUNet-ours  & 33.13/0.900 & \textbf{31.12}/\textbf{0.865} & \textbf{31.86}/\textbf{0.896} & \textbf{31.60}/\textbf{0.883} & \textbf{27.10}/\textbf{0.800} & \textbf{30.91}/\textbf{0.873} & \textbf{32.10}/\textbf{0.896} & \textbf{31.58}/\textbf{0.862}  \\ 
        \hline \hline
        DMID $\sigma=15$ & 24.34/0.485 & 17.41/0.218 & 12.19/0.120 & 15.08/0.207 & 15.07/0.447 & 21.14/0.532 & 23.53/0.655 & 22.29/0.403  \\ 
        DMID-ours & \textbf{33.26}/\textbf{0.902} & \textbf{31.23}/\textbf{0.868} & \textbf{31.06}/\textbf{0.875} & \textbf{31.12}/\textbf{0.869} & \textbf{27.41}/\textbf{0.788} & \textbf{30.69}/\textbf{0.865} & \textbf{31.65}/\textbf{0.883} & \textbf{31.68}/\textbf{0.863}  \\ 
        \bottomrule
    \end{tabular}
\end{table*}

Table \ref{table1} and Table \ref{table2} present quantitative comparison results. The measurement indicators are PSNR and SSIM. For typical Gaussian denoisers such as CBM3D \cite{CBM3D}, DnCNN \cite{DnCNN}, FFDNet \cite{FFDNet}, DRUNet \cite{DRUNet}, PD-denoising \cite{PD}, and Restormer \cite{Restormer}, by our method, their denoising performance improves significantly. For example, our method, utilizing a Gaussian denoiser with $\sigma=15$, achieves performance comparable to that of the supervised Gaussian denoiser with $\sigma=25$ when dealing with Gaussian noise of $\sigma=25$. Note that PD-denoising \cite{PD} performs well on random impulse noise because its training data contains random impulse noise. Generally, the better they remove Gaussian noise, the better the performance of our method based on them will be, but this is not always the case. Restormer-ours exhibits inferior denoising performance compared to FFDNet-ours on synthetic noises (e.g. salt-and-pepper) and real-world noises. This is because flexible Gaussian denoisers are usually more robust than single-level Gaussian denoisers \cite{FFDNet}. This also means that if the Gaussian denoiser is severely overfitted to Gaussian noise, it will lead to the failure of the noise transformation. Therefore, we recommend using a flexible Gaussian denoiser. On the one hand, it is more robust; on the other hand, it allows us to set different target noise levels.

For Clipdenoising \cite{Clipde}, it has trained two models: one for synthetic noise (with Gaussian noise) and one for real-world noise (with Poisson-Gaussian noise). As shown in the Table \ref{table1}, its Gaussian model generalizes poorly to real-world noise, while its Poisson-Gaussian model fails to handle diverse synthetic noises. In contrast, Clipdenoising-ours, using only a Gaussian denoising model, performs well on both synthetic and real-world noises, enhancing generalization. For DMID \cite{DMID}, though capable of noise conversion, has limited ability to convert diverse synthetic noises and requires thousands of iterations—compared to just three for our method. For SCUNet \cite{SCUNet}, its blind denoising model SCUNet-real is trained on mixed noise and performs well on both synthetic noise and real noise. However, it still has the problem of insufficient generalization and cannot remove noise such as salt and pepper noise. SCUNet-ours outperforms SCUNet-real with models trained only with Gaussian noise. For real-world noise removal methods, AP-BSN \cite{AP-BSN}, LTN \cite{LTN}, TBSN \cite{TBSN}, they perform well on real-world noise but poorly on synthetic noise. MASH \cite{MASH} is a zero-shot method, it exhibits good generalization ability, yet mediocre denoising performance.

\subsection{Ablation study}
\subsubsection{Qualitative analysis}
\begin{figure}[htbp]
    \centering
    \includegraphics[width=0.9\columnwidth]{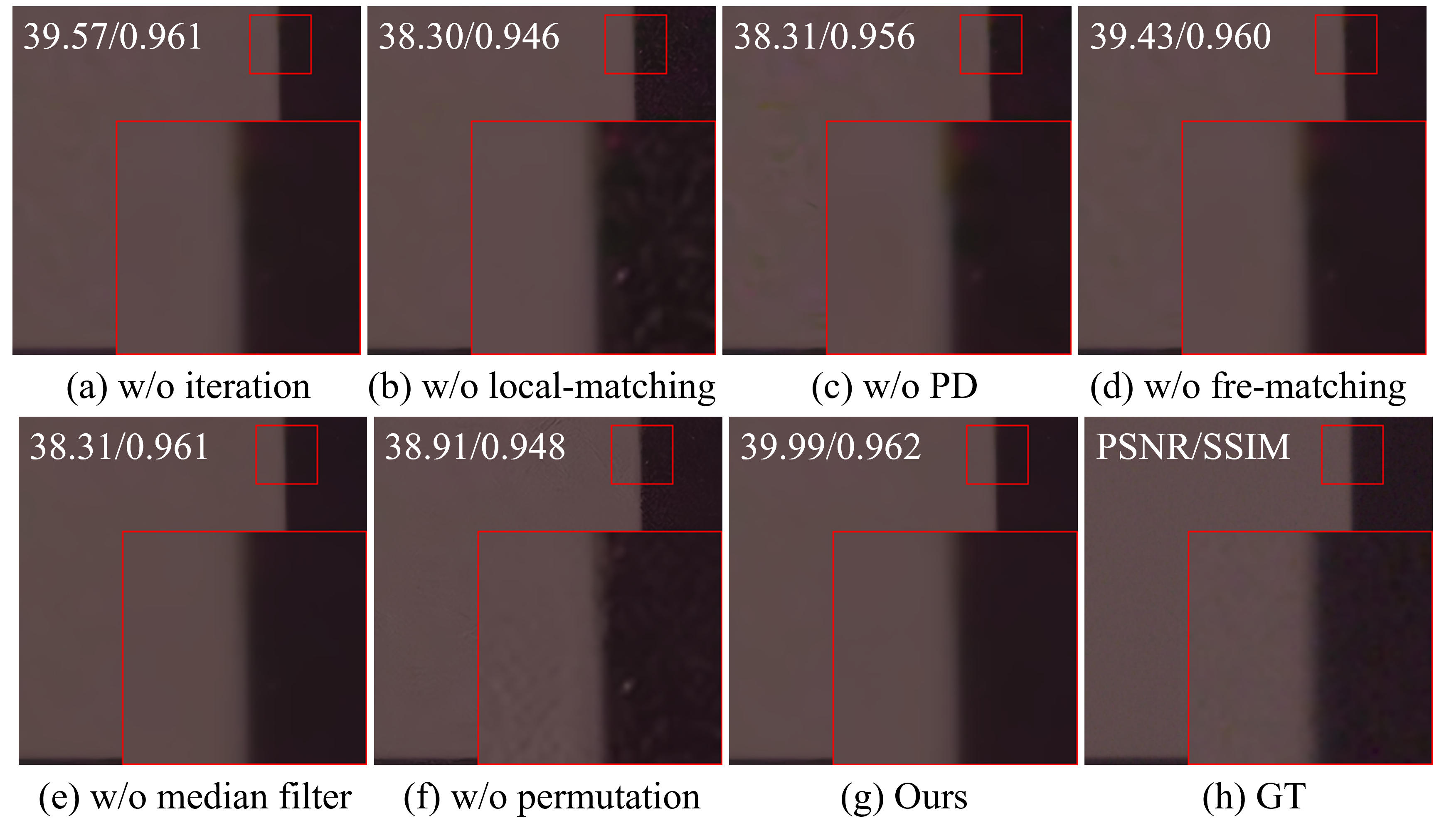}
    \caption{Qualitative analysis of each step in our method on real-world noise.}
    \label{fig5}
\end{figure}

Fig. \ref{fig5} visually presents the effects of each step on the real-world noise. As shown, residual noise or artifacts remain in the absence of local and frequency-domain histogram matching, PD, and intrapatch permutation. Without median filtering, noise is removed, but overall brightness is slightly dim. Without iteration, the denoised result is smooth. Next, we will provide a more specific analysis.

\begin{figure}[htbp]
    \centering
    \includegraphics[width=0.98\columnwidth]{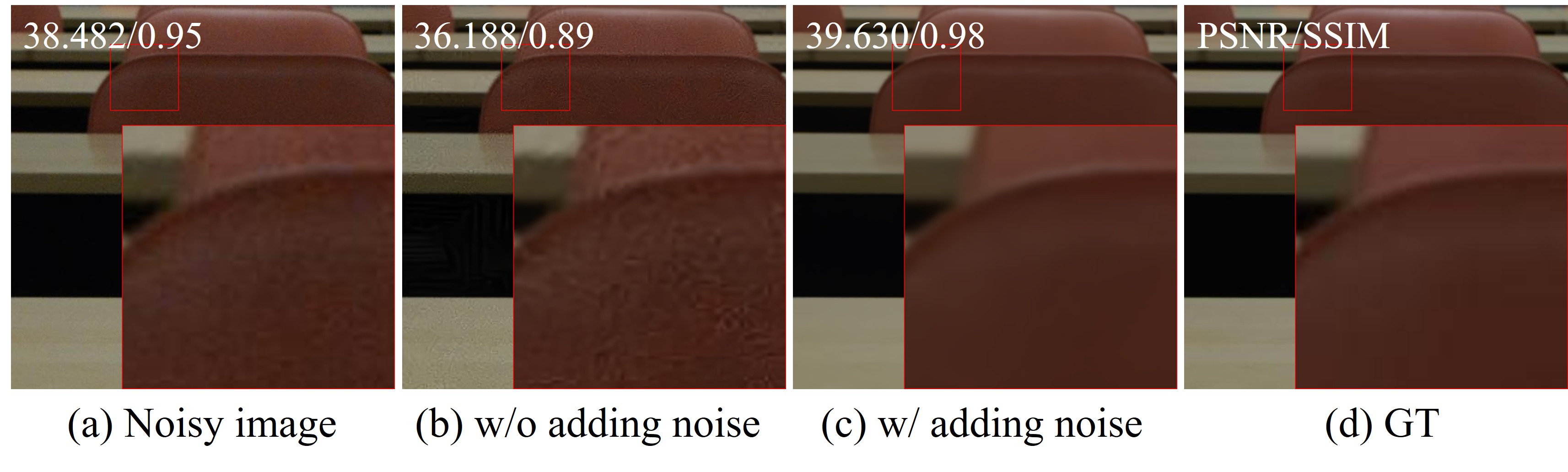}
    \caption{Qualitative analysis on adding noise. The noisy image is real-world noisy image in JPG format. It can be seen that adding noise improves robustness.}
    \label{add}
\end{figure}

\textbf{Analysis on Adding noise}: Adding noise before noise transformation can improve robustness, since the noise level in some scenarios is very low. As shown in Fig. \ref{add}, the noisy image depicted exhibits high PSNR and SSIM values, a low noise level, and also contains JPG compression artifacts. In this case, if no noise is added, the final denoising result will instead magnify the artifacts, leading to a degradation in denoising performance. Conversely, with adding noise, we obtain a good denoising result.

\begin{figure}[htbp]
    \centering
    \includegraphics[width=0.85\columnwidth]{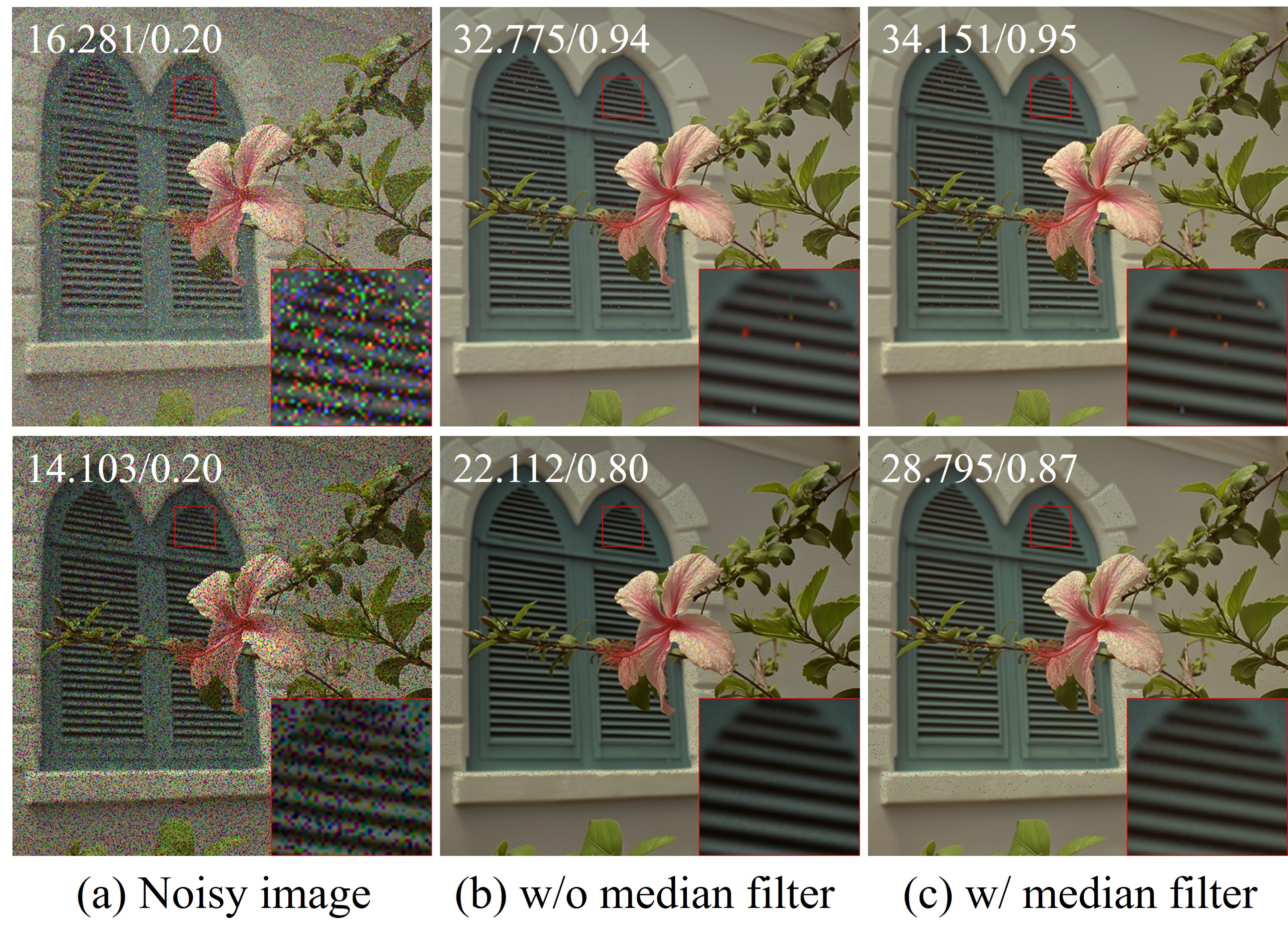}
    \caption{Qualitative analysis on median filter. The first row is random impulse noise and the second row is Bernoulli noise. Without median filter, the PSNR declines significantly.}
    \label{median}
\end{figure}

\textbf{Analysis on Median filter}: The median filter is designed for non-zero mean noise. This is because when noise exhibits a non-zero mean, image smoothing  tends to cause deviations in image brightness, which in turn leads to brightness shifts in the final denoised result. Using median filtering can bring the image brightness back. As shown in the Fig. \ref{median}, adding a simple median filter before smoothing can greatly improve the denoising effect on non-zero mean noise.

\begin{figure}[t]
    \centering
    \includegraphics[width=0.98\columnwidth]{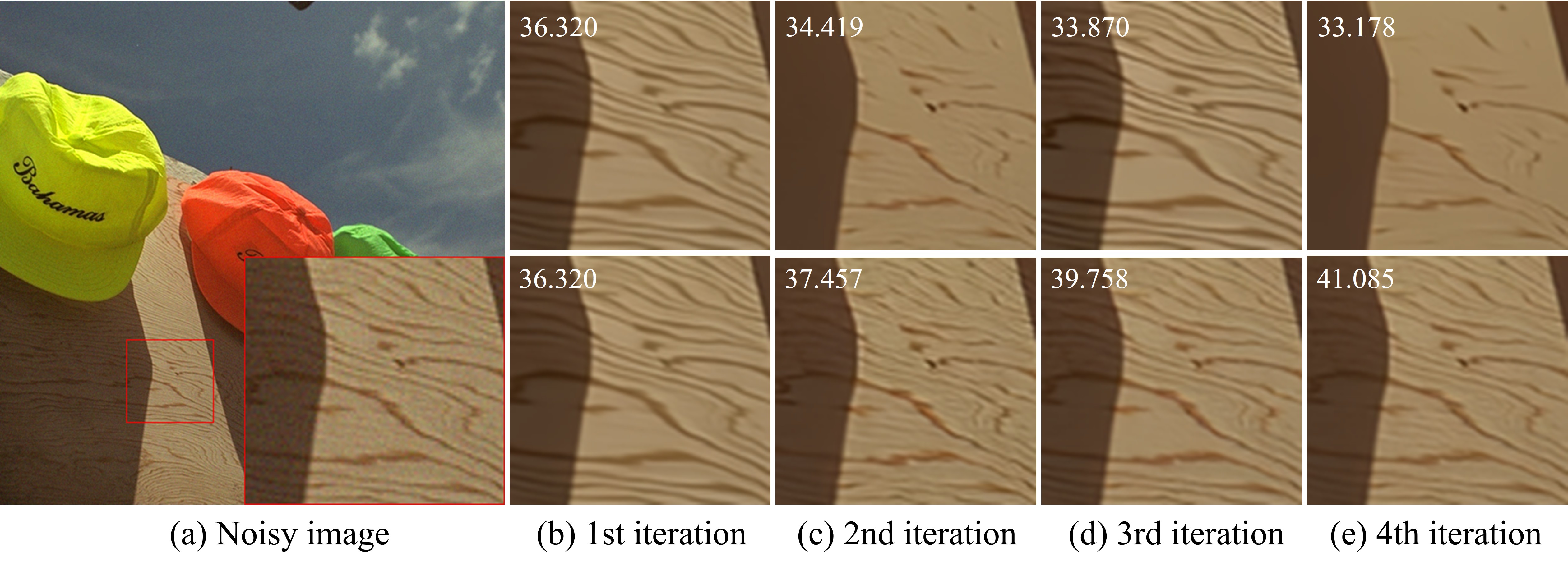}
    \caption{Qualitative analysis on texture transformation. The first row shows the iteration process without texture transformation, and the second row shows the iteration process with texture transformation. Iterations with texture transformation are smoother.}
    \label{texture-trans}
\end{figure}

\textbf{Analysis on Texture transformation}: Texture transformation is designed to ensure the smooth progress of iterations. If without texture transformation, the phenomenon of texture cancellation may occur, as shown in Fig. \ref{texture-trans}. The noise in the figure is Gaussian noise with $\sigma=5$. Due to this low noise level, while the target noise level is 15, our noise transformation operation will enhance some textures in the first iteration. At this time, if the denoising result is directly used for the next round of noise transformation, the phenomenon of texture cancellation will occur. In subsequent iterations, the texture will be enhanced again, but then cancelled out, resulting in the failure of the iteration. However, with texture transformation, the iteration proceeds smoothly, and the denoising effect gradually improves.

\begin{figure}[t]
    \centering
    \includegraphics[width=0.98\columnwidth]{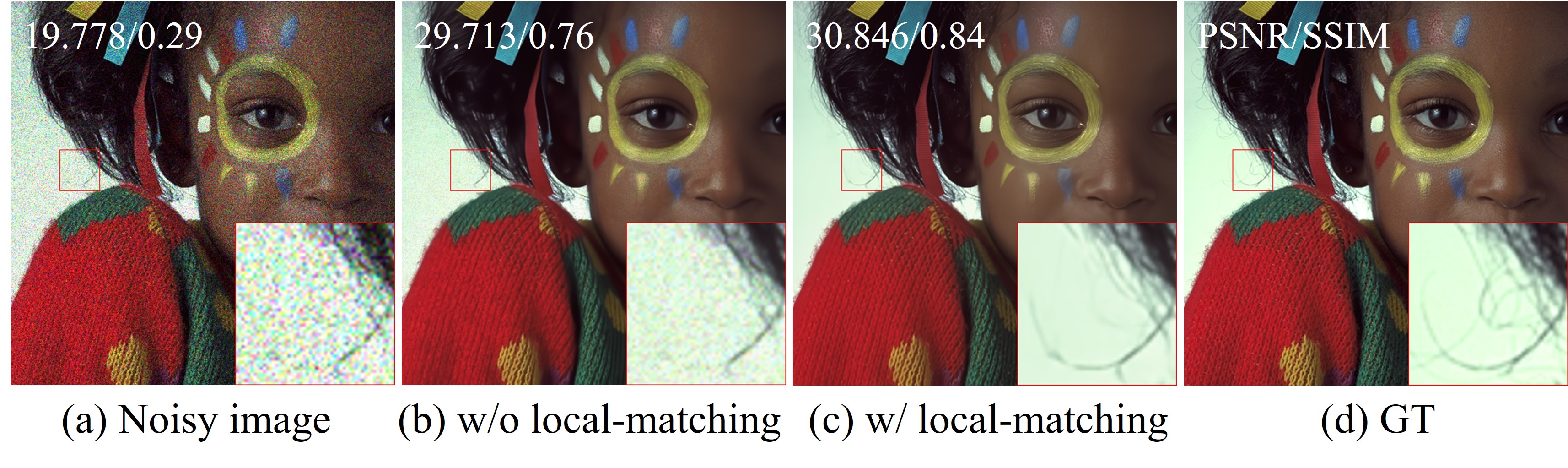}
    \caption{Qualitative analysis on local histogram matching. The noisy image is Poisson noise image. As can be seen, local histogram matching is important for local noise.}
    \label{local-match} 
\end{figure}

\textbf{Analysis on Local histogram matching}: Local histogram matching is very necessary for signal-dependent noise, as Gaussian denoisers are global denoisers. Fig. \ref{local-match} shows the effect of local histogram matching. The noise present in the figure is Poisson noise. In the absence of local histogram matching, there is residual noise. Conversely, with the application of local histogram matching, a favorable denoising effect is achieved.

\begin{figure}[t]
    \centering
    \includegraphics[width=0.98\columnwidth]{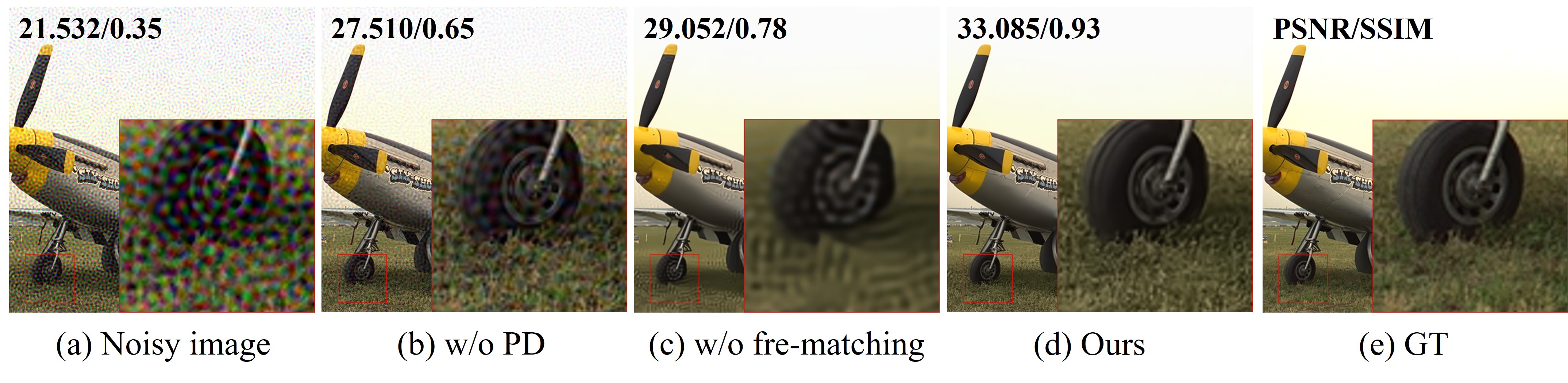}
    \caption{Qualitative analysis on frequency-domian histogram matching. The noisy image is circular repeating pattern noise image. Both PD and frequency-domain histogram matching play a crucial role.}
    \label{fre-match}
\end{figure}

\textbf{Analysis on Frequency-domain histogram matching}: Only pixel-shuffle down-sampling cannot completely break the spatial correlation of noise especially when the noise also has a certain repetitive pattern, so we introduce frequency-domain histogram matching. Fig. \ref{fre-match} shows the effect of both. It can be seen that both PD and frequency-domain histogram matching play crucial roles and are indispensable. Without PD, noise cannot be removed and without frequency-domain histogram matching, artifacts are produced.

\begin{figure}[t]
	\centering
	\includegraphics[width=0.98\columnwidth]{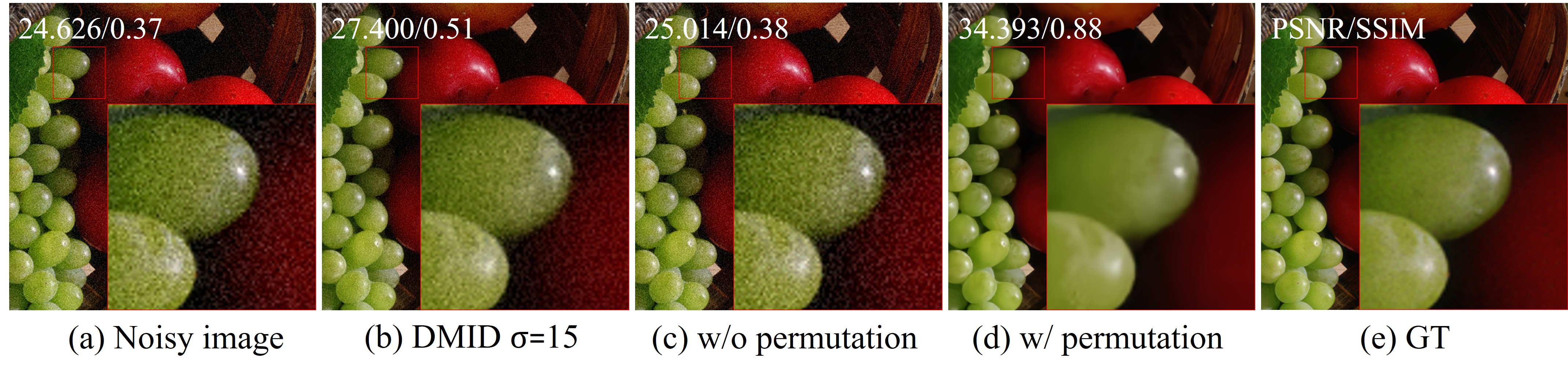}
	\caption{Qualitative analysis on intrapatch permutation. The noisy image is channel-dependent Gaussian noise image. Intrapatch permutation effectively breaks the channel correlation.}
	\label{permu}
\end{figure}

\textbf{Analysis on Intrapatch permutation}: Intrapatch permutation breaks channel correlation, and Fig. \ref{permu} shows its effect. The noise in the figure is channel-dependent noise, with the same Gaussian noise of $\sigma=15$ added to three channels. In this case, due to the channel correlation, the Gaussian denoiser fails to denoise. Without intrapatch permutation, noise still exists. And with intrapatch permutation, we achieve a good denoising result.

\subsubsection{Quantitative analysis}
\frenchspacing
Quantitative analysis of each step is shown in Table \ref{table3}. And the test method is DMID-ours. In the Table \ref{table3}, adding noise seems to only have a little effect, but in fact, it is very useful in some low-level-noise scenarios and can help improve robustness (see the Analysis on Adding noise). Median filtering markedly improves performance on random impulse noise and real-world noise, as both are non-zero-mean noise (real world noisy images may have dead points that cause noise not to be non-zero mean). However, for zero-mean noise, the use of median filtering causes a slight performance degradation. Given the practical difficulty in distinguishing between zero-mean and non-zero-mean noise, the application of median filtering remains beneficial overall. Iteration and texture transformation processes are also critical: the former enhances transformation and denoising effects, while the latter ensures smooth iteration. Without local histogram matching, denoising performance on Poisson noise and real-world noise degrades. For spatially correlated noise, both PD and frequency-domain matching are indispensable. Intrapatch permutation also contributes significantly to removing real-world noise.

\begin{table}[t]
	\caption{Quantitative analysis of each step in our method. Due to the space limitation, the table shows the abbreviations.}
	\label{table3}
	\setlength{\tabcolsep}{1mm}
	\renewcommand\arraystretch{1}
	\small
	\centering
	\begin{tabular}{cccccc}
		\toprule
		Noise Types& \makecell[c]{Gaussian\\$\sigma=25$} & \makecell[c]{Impulse\\$d=0.2$} & \makecell[c]{Poisson\\$\lambda=25$} & \makecell[c]{Circular\\$\sigma=25$} & \makecell[c]{Real\\noise}
		\\ \hline \hline
		w/o add-noise & 32.85 & 31.56 & 30.47 & 31.89 & 36.09  \\ 
		w/o median-filt & \textbf{32.89} & 29.73 & \textbf{30.89} & \textbf{32.24} & 35.47  \\ 
		w/o iteration & 28.05 & 27.13 & 26.44 & 26.68 & 35.39  \\ 
		w/o texture-trans & 32.30 & 29.85 & 29.28 & 31.18 & \underline{36.11}  \\ 
		w/o local-match & / & / & 29.99 & / & 36.08  \\ 
		w/o PD & / & / & / & 26.70 & 35.15  \\ 
		w/o fre-match & / & / & / & 29.05 & 35.92  \\ 
		w/o intra-permu & / & / & / & / & 34.95  \\ 
		ours & 32.83 & \textbf{31.58} & \underline{30.49} & \underline{32.13} & \textbf{36.20}  \\ 
		\bottomrule
	\end{tabular}
\end{table}

\begin{figure}[htbp]
	\centering
	\includegraphics[width=0.9\columnwidth]{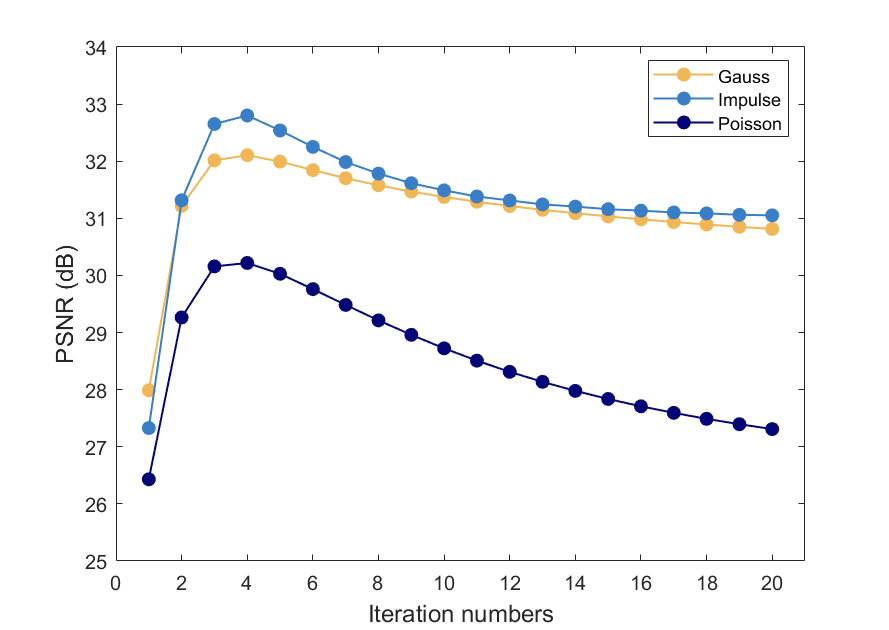}
	\caption{Denoising performance under different iteration numbers.}
	\label{fig6}
\end{figure}

\textbf{Analysis on the number of iterations}: Fig. \ref{fig6} shows the role of iteration and explains why we set the number of iterations to 3. It can be seen that the PSNR value gradually increases in the first four iterations, but the increase is very small in the fourth iteration. Taking into account both time and performance  comprehensively, we set the iteration number to 3. When the iteration number is greater than 4, the PSNR value begins to decrease. This is reasonable because denoising is limited, as is noise transformation. When denoising performance begins to deteriorate, noise transformation performance also starts to decline, resulting in a sustained downward trend.

\subsection{More visualization results}
\begin{figure}[t]
    \centering
    \includegraphics[width=0.95\columnwidth]{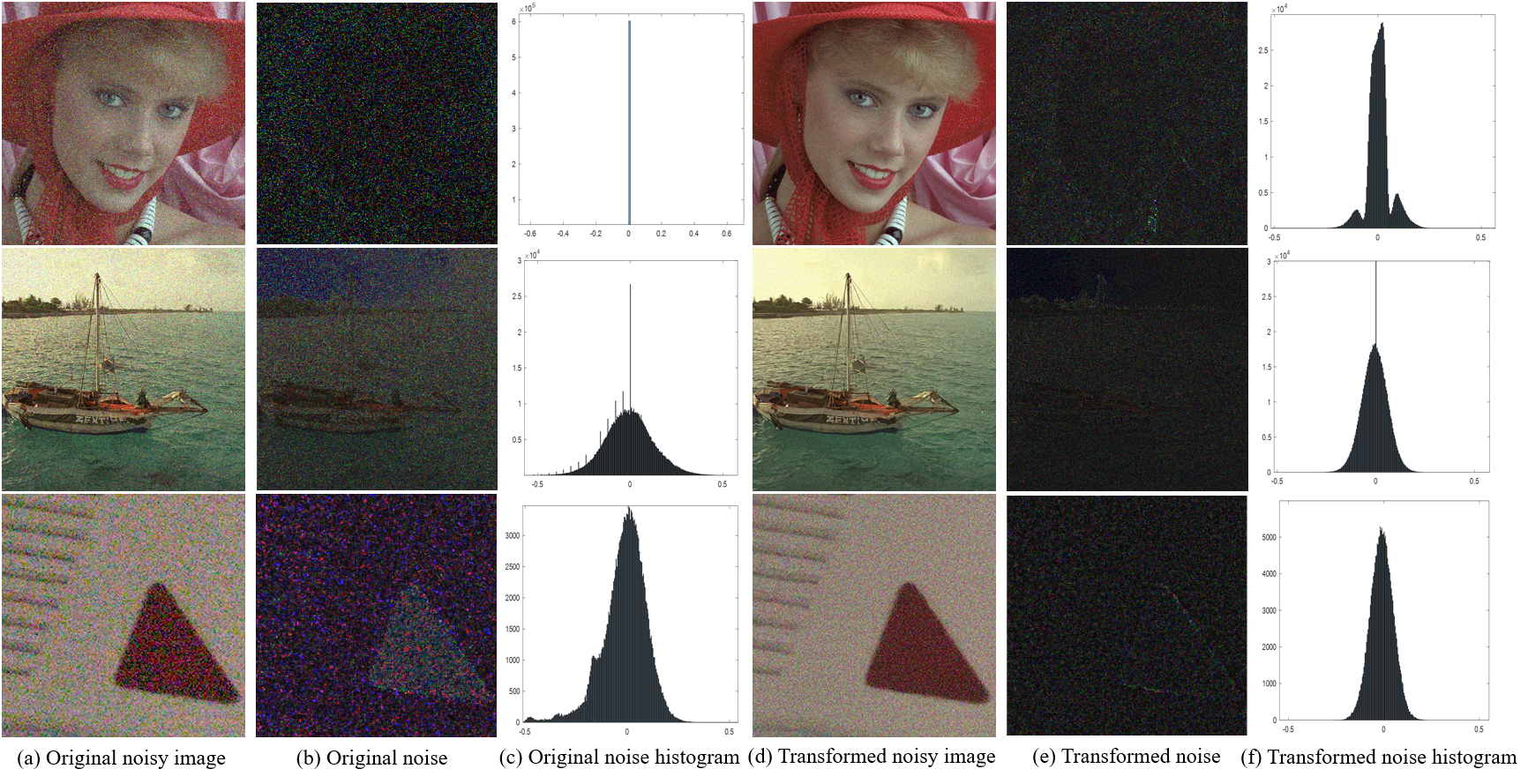}
    \caption{Demonstration of noise transformation effect. The noisy images from the first row to the last row are random impulse noise, Poisson noise, and real-world noise, respectively.}
    \label{noise_trans}
\end{figure}

\begin{figure}[t]
	\centering
	\includegraphics[width=0.98\columnwidth]{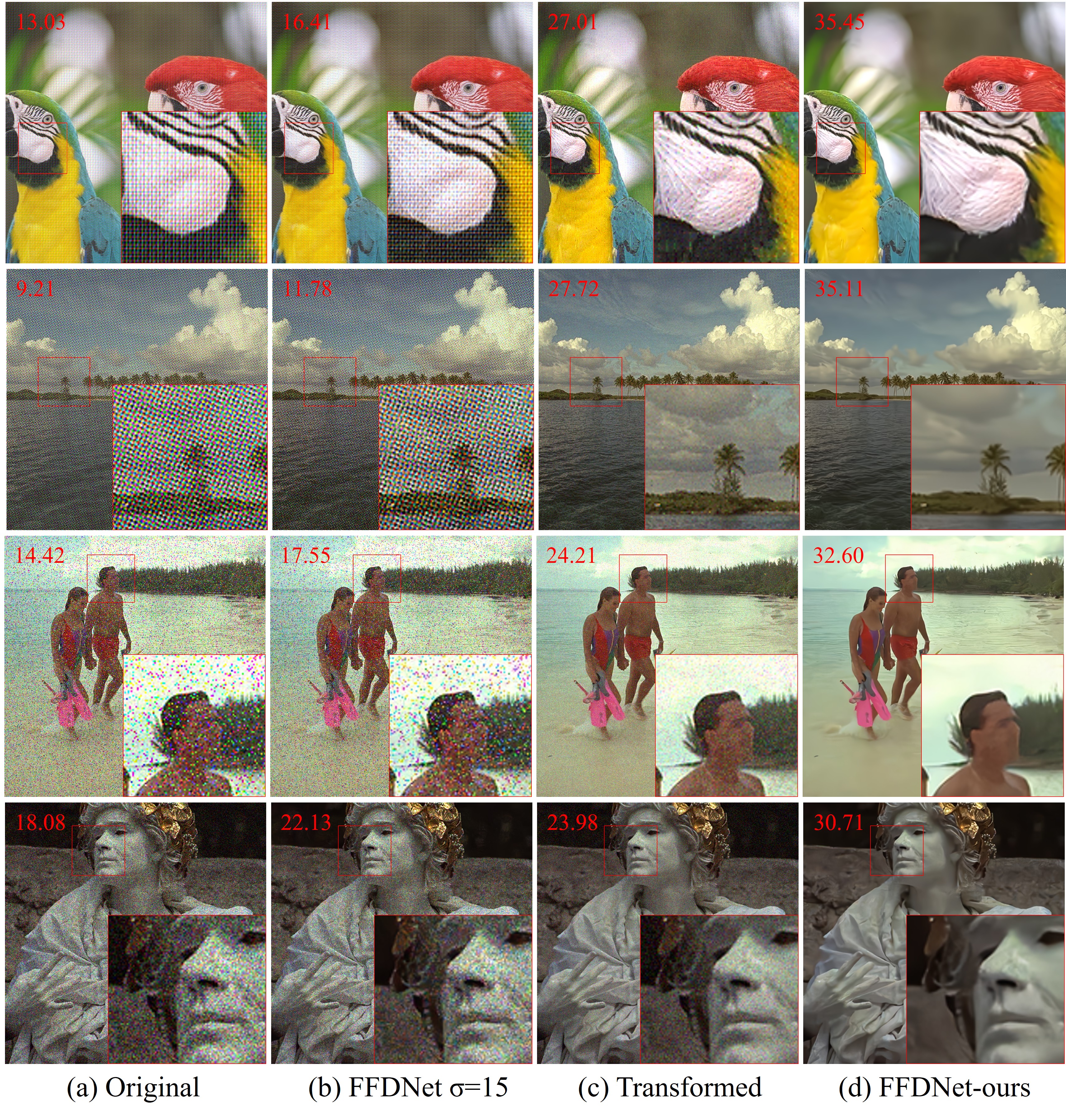}
	\caption{Conversion effects of more types of noise. The noise images from the first row to the last row are stripe noise, grid noise, Gauss-salt-pepper mixed noise and Gauss-Poisson mixed noise respectively.}
	\label{More}
\end{figure}

Fig. \ref{noise_trans} visually demonstrates the effect of noise transformation.As can be seen from the figure, the original noise distribution varies, while the transformed noise distributions tend to be similar to a Gaussian distribution. Fig. \ref{More} shows the transformation effect of our method on more types of noise. Among them, the stripe noise is obtained by adding stripes of different angles to the RGB three channels, while the grid noise is achieved by adding two different angles of stripe noise to each channel, and the angles of three channels are also different. They are all periodic noises, and Gaussian denoisers generally cannot remove them. However, after our noise transformation, we have achieved very good denoising results. Gaussian salt-and-pepper noise and Gaussian Poisson noise are both mixed noises, and our method is also applicable.

\section{Conclusion}
In this paper, we focus on the generalization problem of existing Gaussian denoisers, introducing a histogram matching method to transform noise. Specifically, we design global/local histogram matching and frequency-domain histogram matching strategies for noise transformation. Meanwhile, we employ pixel-shuffle downsampling and intra-patch permutation to disrupt the spatial and channel correlations of noise. To enhance the conversion and denoising effects, we also establish an iterative process between denoising and noise transformation. The proposed method significantly improves the denoising capability of Gaussian denoisers. However, our method requires determining noise properties, which can be challenging. How to decide the level of the target noise is also worth discussing. Moreover, our method is an approximate one, so the noise level after transformation will be close to the expected noise level, but there will be errors. Combining it with a noise level estimator may achieve better results. In future work, we consider further optimizing our approach by introducing a method for determining noise properties and adaptively selecting the target noise level.


\bibliographystyle{IEEEtran.bst}
\bibliography{my.bib}

\end{document}